\documentclass{article}

\makeatletter
\@ifundefined{ifanon}{\newif\ifanon}{}
\makeatother

\ifanon
  \usepackage[nonatbib]{neurips_2026}
\else
  \usepackage[preprint,nonatbib]{neurips_2026}
\fi

\usepackage[utf8]{inputenc}
\usepackage[T1]{fontenc}

\usepackage{hyperref}
\usepackage{url}
\usepackage{xcolor}
\usepackage{amsmath}
\usepackage{amssymb}
\usepackage{booktabs}
\usepackage{longtable}
\usepackage{array}
\usepackage{calc}
\providecommand{\real}[1]{#1}
\usepackage{caption}
\usepackage{enumitem}
\usepackage{verbatim}
\usepackage{tikz}
\usetikzlibrary{positioning,arrows.meta,shapes.geometric,calc,fit,backgrounds}

\usepackage{newunicodechar}
\newunicodechar{≈}{\ensuremath{\approx}}
\newunicodechar{≠}{\ensuremath{\neq}}
\newunicodechar{×}{\ensuremath{\times}}
\newunicodechar{−}{\ensuremath{-}}
\newunicodechar{∈}{\ensuremath{\in}}
\newunicodechar{‖}{\ensuremath{\Vert}}
\newunicodechar{→}{\ensuremath{\rightarrow}}
\newunicodechar{…}{\ensuremath{\ldots}}
\newunicodechar{‑}{-}
\newunicodechar{⁰}{\textsuperscript{0}}
\newunicodechar{¹}{\textsuperscript{1}}
\newunicodechar{²}{\textsuperscript{2}}
\newunicodechar{³}{\textsuperscript{3}}
\newunicodechar{⁴}{\textsuperscript{4}}
\newunicodechar{⁵}{\textsuperscript{5}}
\newunicodechar{⁶}{\textsuperscript{6}}
\newunicodechar{⁷}{\textsuperscript{7}}
\newunicodechar{⁸}{\textsuperscript{8}}
\newunicodechar{⁹}{\textsuperscript{9}}
\newunicodechar{⁻}{\textsuperscript{\ensuremath{-}}}
\newunicodechar{§}{\S}

\makeatletter
\@ifundefined{c@none}{\newcounter{none}}{}
\makeatother

\providecommand{\tightlist}{\setlength{\itemsep}{0pt}\setlength{\parskip}{0pt}}

\usepackage{framed}
\definecolor{shadecolor}{RGB}{248,248,248}

\hypersetup{
  colorlinks=true,
  linkcolor=blue,
  urlcolor=blue,
  citecolor=blue,
}

\title{Sutra: Tensor-Op RNNs as a Compilation Target\\
  for Vector Symbolic Architectures}

\ifanon
  \author{Anonymous Authors}
\else
  \author{%
    Emma Leonhart\\
    \texttt{emma@topazcomputing.com}%
  }
\fi

\begin{document}
\maketitle

\begin{center}\rule{0.5\linewidth}{0.5pt}\end{center}

\hypertarget{abstract}{%
\section*{Abstract}\label{abstract}}
\addcontentsline{toc}{section}{Abstract}

\textbf{Sutra} is a typed, purely functional programming language whose compiled forward pass is a PyTorch neural network. The compiler beta-reduces the whole program --- primitives, control flow, string I/O --- to a single substrate-pure tensor-op dataflow graph over a frozen embedding substrate (every operation is a tensor op; the language has no scalar-readout escape hatch). Rotation binding, unbind, bundle, polynomial Kleene three-valued logic, and tail-recursive loops all lower to tensor operations; the Kleene connectives are Lagrange-interpolated polynomials exact on the \{−1, 0, +1\} truth grid.

Validation is one fact tested two ways. (1) The same program runs on four frozen embeddings spanning two modalities --- three text encoders (nomic-embed-text, all-minilm, mxbai-embed-large) and one protein language model (ESM-2) --- and decodes bundles at 100\% accuracy through width k=8 on every substrate, where the textbook Hadamard product has already collapsed (2.5\% on mxbai-embed-large, 7.5\% on all-minilm). (2) PyTorch autograd flows through the actually compiled graph: a fuzzy-rule classifier written in \texttt{.su} trains from random init (18.7 ± 9.5\%; chance = 20\%, five classes) to 100.0 ± 0.0\% (three seeds) by backpropagating through the emitted graph, the symbolic source unmodified. A weighted variant additionally trains a scalar cosine gain and writes it back into the \texttt{.su} source as a numeric literal; recompiling reproduces the trained behaviour to ≈ 2×10⁻⁷ per logit, so the trained model is itself legible, recompilable code.

The same artifact is therefore both a logic program and a trainable neural network.

\begin{center}\rule{0.5\linewidth}{0.5pt}\end{center}

\hypertarget{introduction}{%
\section{Introduction}\label{introduction}}

A frozen embedding model maps strings (or amino-acid sequences,
or any other input the model was trained on) into a
deterministic continuous vector space. Given such a substrate,
two technical questions follow:

\begin{enumerate}
\def\labelenumi{\arabic{enumi}.}
\tightlist
\item
  \textbf{Which operations on these embeddings are reliable enough to
  be used as primitives} of a compositional algebra over the
  substrate\textquotesingle s vector space?
\item
  \textbf{What is the correct binding operation?} Hyperdimensional
  computing\textquotesingle s textbook bind operators (Hadamard product,
  circular convolution) were derived assuming hypervectors
  drawn from a controlled random distribution. Frozen LLM
  embeddings are not such a distribution: they are strongly
  \emph{anisotropic}, concentrating in a narrow cone so cosine
  similarity is compressed and inflated even between unrelated
  items (Ethayarajh 2019). §3.2 measures four
  substrates and reports that rotation binding decodes at 100\%
  accuracy through bundle widths where Hadamard has already
  collapsed.
\end{enumerate}

This paper answers both questions in the form of a working
programming language, \textbf{Sutra}, whose primitives are these
consolidated operations and whose compiled forward pass is a
PyTorch neural network. The naming: \textbf{Sutra} is the Sanskrit
\emph{sūtra} (thread, rule, aphorism), the term for Pāṇini\textquotesingle s
foundational Sanskrit grammar.

\hypertarget{contributions}{%
\subsection{Contributions}\label{contributions}}

The four core technical contributions of this paper are:

\begin{enumerate}
\def\labelenumi{\arabic{enumi}.}
\item
  \textbf{Polynomial fuzzy logic via Lagrange interpolation of
  Kleene\textquotesingle s three-valued truth tables.} The truth axis encodes
  \(T = +1\), \(U = 0\), \(F = -1\). On the discrete
  \(\{-1, 0, +1\}\) grid, the Kleene connectives are
  \(\mathrm{AND} = \min\), \(\mathrm{OR} = \max\), \(\mathrm{NOT} = -\,\cdot\,\).
  The min/max forms (the standard Gödel t-norm/t-conorm choice;
  Hájek 1998) are non-differentiable at the diagonal \(a = b\),
  which breaks gradient flow when connectives compose with the
  tensor-op graph (van Krieken, Acar \& van Harmelen 2022 survey
  the issue across t-norm-derived neural-symbolic operators).
  Sutra resolves this by Lagrange-interpolating each connective
  as a polynomial that is exact on the \(3\times 3\) Kleene grid
  and \(C^{\infty}\) elsewhere:

  \begin{align*}
  \mathrm{AND}(a, b)  &= \tfrac{1}{2}(a + b + ab - a^2 - b^2 + a^2 b^2) \\
  \mathrm{NAND}(a, b) &= \tfrac{1}{2}(-a - b - ab + a^2 + b^2 - a^2 b^2) \\
  \mathrm{OR}(a, b)   &= \tfrac{1}{2}(a + b - ab + a^2 + b^2 - a^2 b^2) \\
  \mathrm{NOR}(a, b)  &= \tfrac{1}{2}(-a - b + ab - a^2 - b^2 + a^2 b^2) \\
  \mathrm{NOT}(a)     &= -a \\
  \mathrm{XOR}(a, b)  &= -ab \\
  \mathrm{XNOR}(a, b) &= ab
  \end{align*}

  \{AND, OR, NOT\} is functionally complete for the Kleene
  fragment; \(\mathrm{NAND}\) and \(\mathrm{NOR}\) are just \(-\mathrm{AND}\) and \(-\mathrm{OR}\) (negation is \(-\,\cdot\,\)), and XOR/XNOR collapse to a single multiplicative term
  because their interpolant is zero whenever either input is U
  and bilinear in the \{−1, +1\} corners. Every Kleene-valid
  connective is therefore a polynomial tensor-op-graph fragment,
  gradient-compatible, branchless, and exact on the
  discrete-logic regime. A symbolic if-then rule built from
  these gates is one fused subgraph that PyTorch autograd
  backprops through end-to-end (§3.6).

  The choice of Lagrange interpolation over softer alternatives
  --- softened t-norms such as the Einstein or Yager families,
  or soft-min/soft-max with a temperature --- is fixed by a
  non-negotiable requirement: agreement with classical Kleene
  reasoning on the discrete grid must be bit-for-bit, not
  approximate. Softened t-norms drift from \(\{-1, 0, +1\}\)\textquotesingle s
  truth values at the grid corners themselves; Lagrange
  interpolation on the \(3 \times 3\) Kleene grid is the unique
  polynomial that \emph{exactly} recovers each connective at the
  nine grid points while remaining \(C^{\infty}\) off it. The
  trade-off is one of monotonicity: the bilinear-\/-quadratic
  interpolant is non-monotonic at some interior points (a
  defuzzified boolean that happens to read \(a = 0.5,\ b = 0\)
  produces \(\mathrm{AND}(a, b) = 0.125\), peaking off-grid before
  dropping back to \(0\) at \(a = 1\)). Sutra accepts this in
  exchange for grid-exactness; replacing the interpolant with a
  higher-degree monotonic polynomial is a future direction.
\item
  \textbf{Beta reduction to a substrate-pure tensor-op graph.}
  The compiler inlines stdlib operator definitions,
  beta-reduces through bound names, then runs an
  algebraic-simplification pass over the residual. What\textquotesingle s
  left is a fused tensor-op graph (matmul / element-wise /
  nonlinear) with no named bindings or function calls. Three concrete moves go beyond standard
  inlining + constant folding: conditionals lower to soft-mux
  polynomials (\(\tfrac{1+\mathrm{cond}}{2}\,a + \tfrac{1-\mathrm{cond}}{2}\,b\)) so the compiled
  artifact has no \texttt{if} opcodes; Haar-orthogonal binding
  rotations \texttt{R\_role} are materialized at compile time so
  runtime \texttt{bind} is one matmul against a constant matrix;
  canonical synthetic axes are assigned compile-time so every
  primitive-type read/write is a known index, not a hashtable
  lookup. §4.2 traces this lowering stage-by-stage on a
  concrete program; the compilation pipeline as a whole is
  diagrammed in Figure\textasciitilde{}\ref{fig:compile-pipeline}.
\item
  \textbf{Tail recursion as the loop primitive.} Loops are
  tail-recursive function declarations (\texttt{do\_while},
  \texttt{while\_loop}, \texttt{iterative\_loop}, \texttt{foreach\_loop}) whose body\textquotesingle s
  \texttt{return\ NAME(args)} becomes the recurrent step. Each loop
  compiles to a soft-halt RNN cell with substrate-pure halt
  detection (heaviside → cumulative monotone halt → soft-mux
  state freeze). The body of every loop tick is one
  straight-line tensor pipeline with no in-graph branches; a
  thin Python \texttt{while\ True:\ …\ break} driver wraps the body and
  terminates when the halt scalar saturates (§3.4). The state
  vector is fixed-width across iterations, \textbf{O(1) state, O(N)
  compute, O(N) gradient tape during training}, where N is
  iterations actually executed.
\item
  \textbf{Synthetic-dimension rotation binding as an angular hash map.}
  The compiler reserves a synthetic block of canonical
  dimensions and uses Haar-orthogonal rotations seeded from the
  role\textquotesingle s content hash to bind keys to slots. We are unaware of
  any prior use of a high-dimensional rotation pattern as the
  substrate for a functional hash-map primitive.
\end{enumerate}

These four primitives integrate into a single working compiler
that lowers \texttt{.su} source to a self-contained PyTorch module on
CPU or CUDA. Program inputs and outputs are embeddings in the
substrate\textquotesingle s vector space; a compile-time codebook (implemented
with an embedded vector database, §3.5) handles the convenience
of source-level string literals and nearest-string output.

\hypertarget{the-substrate-is-the-architecture-target}{%
\subsection{The substrate is the architecture target}\label{the-substrate-is-the-architecture-target}}

A Sutra program is compiled for an \emph{embedding-space architecture},
the way a C program is compiled for x86 and a CUDA kernel for an
NVIDIA SM. The embedding model fixes dimensionality, the geometry
of the semantic block, and the meaning of every basis-vector
lookup; swap the model and the same source recompiles to a
different \texttt{.sdb} codebook against a different geometry. The
substrate need not be an LLM, it can be any network producing a
dense vector representation, including the hidden state of a
trained model. §3.2\textquotesingle s ESM-2 protein-LM row demonstrates this
substrate-agnostically.

\begin{center}\rule{0.5\linewidth}{0.5pt}\end{center}

\hypertarget{related-work}{%
\section{Related Work}\label{related-work}}

\hypertarget{vector-symbolic-architectures}{%
\subsection{Vector Symbolic Architectures}\label{vector-symbolic-architectures}}

VSA is a family of algebraic frameworks for computing with high-
dimensional vectors (Kanerva 2009; Plate 1995; Gayler 2003). The
standard VSA development assumes hypervectors drawn from a
controlled random distribution designed for the algebra; bind is
typically Hadamard product or circular convolution. Frozen LLM
embedding spaces are not designed for VSA: they are anisotropic
(Ethayarajh 2019; Gao et al. 2019; Mu et al. 2018) ---
representations concentrate in a narrow cone, so cosine
similarity has low dynamic range and the textbook bind operations
do not transfer cleanly. The same anisotropy makes an unweighted
cosine read a weak rule signal, which is why §3.6 trains the
embeddings the rules compare against rather than relying on raw
cosine. Rotation
binding (\texttt{R\_role\ @\ filler} for a role-seeded Haar-random
orthogonal \texttt{R\_role}) is the choice that worked across the
substrates we tested, and is what Sutra uses today; §3.2
reports the per-substrate measurements supporting that choice.

Recent work gives VSAs a category-theoretic foundation: Shaw et al.
(2025) formalize the binding, bundling, and similarity operations
within a categorical framework and, discussing open directions,
identify the construction of a \emph{purely functional} VSA --- functional
in the programming-language sense --- as a significant engineering
goal. Sutra can be read as an independent, engineering-first
realization of that direction: a functional language whose
primitives are VSA operations, compiled to a differentiable
neural-network forward pass. We arrived at the construction before
encountering their formalization; that two routes --- categorical
axiomatization and engineering-first iteration --- converge on the
same object is some evidence the direction is structurally natural.

The closest software peer in the VSA space is \textbf{TorchHD}
(Heddes et al. 2023), a PyTorch library that exposes VSA
primitives (bind, bundle, similarity) as tensor operations.
Sutra and TorchHD differ on what the user writes and what the
compiler does:

\begin{itemize}
\tightlist
\item
  \textbf{TorchHD is a \emph{library}.} The user writes Python code that
  calls TorchHD primitives; control flow is host-side Python;
  there is no source-language layer above the primitives, no
  compile step, and no algebraic reduction across primitive
  calls. Each primitive call is a tensor op, but the program
  itself is a Python function with whatever control flow the
  user wrote.
\item
  \textbf{Sutra is a \emph{language with a compiler}.} The user writes
  \texttt{.su} source which the compiler beta-reduces to a
  substrate-pure tensor-op graph (§1.1-2): a single
  straight-line graph of matmul / element-wise / nonlinear ops
  with no Python control flow. Loops are tail-recursive function
  declarations that lower to soft-halt RNN cells; conditionals
  are differentiable fuzzy interpolations rather than Python
  \texttt{if}. Hash-map structure is implemented via synthetic-dimension
  rotation, not via a host-side dictionary.
\end{itemize}

A second axis where Sutra differs from existing HDC software is
\textbf{string I/O}. TorchHD and similar libraries expose the algebra
over user-supplied hypervectors; the user maintains a
\texttt{dict{[}str,\ hypervector{]}} and an explicit codebook tensor by hand.
Sutra\textquotesingle s compile-time codebook (§3.5) closes that loop: every
embedded string in \texttt{.su} source is embedded once at compile time
via the configured frozen LLM, stored in the project\textquotesingle s \texttt{.sdb}
codebook, and decoded at the program output via \texttt{nearest\_string}.
The frozen-LLM embedding is load-bearing, random hypervectors
yield a working VSA algebra with no I/O story.

The structural differences (Sutra contains no Python, the
string-to-vector map and codebook are constructed by the compiler
rather than by the user, and the whole program reduces to a single
fused tensor-op graph) are differences in artifact shape, not
library speed.

\hypertarget{comparison-to-other-neuro-symbolic-languages}{%
\subsection{Comparison to other neuro-symbolic languages}\label{comparison-to-other-neuro-symbolic-languages}}

The closest neuro-symbolic-language peers are \textbf{Scallop} (Li et
al. 2023, Datalog with provenance-semiring differentiability),
\textbf{DeepProbLog} (Manhaeve et al. 2018, ProbLog with neural
predicates), \textbf{Logic Tensor Networks} (Badreddine et al. 2022,
first-order logic compiled to t-norm losses), and \textbf{NeurASP}
(Yang et al. 2020, Answer Set Programming with neural predicates).
All share a two-stage perception-then-reasoning shape: a
neural model extracts discrete symbols from raw input, and a
symbolic program reasons over those symbols. Sutra\textquotesingle s shape is
different at this architectural level: the substrate is a
continuous embedding space throughout, primitives operate on
vectors end-to-end, and the whole program (including what would
be the logic program in Scallop) compiles to a single fused
tensor-op graph through beta reduction. There is no discrete
symbolic stratum to extract into or reason over; differentiability
is inherited from the tensor-op graph itself, not from a
provenance annotation on a relational query. The two are good at
different problem structures: Scallop and its peers when the
problem is naturally relational and perception cleanly factors
out; Sutra when computation is best expressed as algebra on
vectors over a substrate the program reads strings into and
decodes strings out of.

The closest HDC peer with compiler infrastructure is \textbf{HDCC}
(Vergés et al. 2023), a description-file DSL targeting
self-contained C for embedded classification, random/level
hypervectors only, no general control flow, scoped to
classification. \textbf{TorchHD} and OpenHD / HDTorch are libraries
without a language-level loop primitive. To the authors\textquotesingle{}
knowledge (literature reviewed through early 2026), no published HDC system combines (a) one fused
tensor-op graph as compile target, (b) HDC primitives as the
operations, (c) a frozen externally-trained vector embedding
space as the substrate, and (d) tail-recursive loops compiled to
soft-halt RNN cells with constant state-vector width in
recursion depth. The combination is what distinguishes Sutra,
not any one of those properties in isolation.

\hypertarget{fuzzy-logic-and-neuro-fuzzy-systems}{%
\subsection{Fuzzy logic and neuro-fuzzy systems}\label{fuzzy-logic-and-neuro-fuzzy-systems}}

Sutra\textquotesingle s polynomial connectives sit downstream of fuzzy set theory
(Zadeh 1965) and the neuro-fuzzy lineage that makes fuzzy inference
trainable --- adaptive-network fuzzy inference systems (Jang 1993)
and the broader fuzzy-neural-network family (Buckley \& Hayashi
1994). Those systems \emph{learn} the fuzzy logic itself: membership
functions and rule parameters are the trainable object. Sutra
inverts this. Its connectives are fixed Lagrange interpolants of
Kleene\textquotesingle s three-valued tables (§1.1-1), exact on the discrete grid
and never tuned; the only learnable parameters are the embeddings
the frozen rule graph evaluates against (§3.6). Sutra is therefore
closer in spirit to the differentiable t-norm analysis of van
Krieken, Acar \& van Harmelen (2022) than to membership-function
learning, and differs from both by compiling the connectives into
a single substrate-pure tensor-op graph rather than evaluating
them in a host interpreter.

\hypertarget{differentiable-programming-aot-compilation-and-knowledge}{%
\subsection{Differentiable Programming, AOT Compilation, and Knowledge}\label{differentiable-programming-aot-compilation-and-knowledge}}

Compilation

The closest design ancestors are partial-evaluation systems that
specialize programs at compile time (the Futamura projections;
Futamura 1971),
differentiable programming systems that treat programs as
differentiable functions (JAX), AOT compilation of neural networks
(TVM, XLA), and knowledge compilation in symbolic AI (Darwiche \&
Marquis 2002). Sutra differs from each: TVM/XLA start from a
network, not toward one; JAX treats programs as differentiable but
does not bake source literals into weights; partial evaluation
specializes for compile-time-known values but does not target a
neural-network-shaped artifact; knowledge compilation targets
Boolean circuits, not continuous embedding spaces. Sutra\textquotesingle s
combination (fold source literals into the weight structure,
compile control flow to RNN cells, run the whole program as one
tensor-op graph over a \emph{continuous} substrate) is the novel
position.

\begin{center}\rule{0.5\linewidth}{0.5pt}\end{center}

\hypertarget{preliminaries}{%
\section{Preliminaries}\label{preliminaries}}

This section consolidates the two prerequisites a reader needs before §3: the
vector-symbolic primitives Sutra compiles to, and the three-valued logic its
conditionals and connectives are built on. Both are used heavily from §3 onward. The
anisotropy of frozen embedding spaces --- the other recurring fact --- is introduced
inline where it bears (§2.1, §3.2, §3.7) rather than here.

\hypertarget{vector-symbolic-primitives}{%
\subsection{Vector-symbolic primitives}\label{vector-symbolic-primitives}}

Sutra\textquotesingle s data-structure primitives come from Vector Symbolic Architectures
(hyperdimensional computing; Plate 1995; Gayler 2003; Kanerva 2009). Four operations
carry the algebra. \emph{Binding} composes a role with a filler into a single vector
dissimilar to both; \emph{unbinding} is its inverse, recovering the filler given the role.
\emph{Bundling} superposes several vectors into one that stays similar to each.
\emph{Similarity} (cosine) measures how close two vectors are, and \emph{cleanup} uses it to
decode a noisy or superposed vector back to the nearest stored item. Together these
let a record, a sequence, or a graph be carried as one fixed-width vector and queried
by vector arithmetic --- the property Sutra\textquotesingle s compiler builds data structures on. §3.1
and Appendix A give the exact operators Sutra uses: rotation binding (\texttt{R\_role\ @\ filler} for a role-seeded Haar-orthogonal \texttt{R\_role}), additive bundling, and cosine
similarity.

\hypertarget{kleene-three-valued-logic}{%
\subsection{Kleene three-valued logic}\label{kleene-three-valued-logic}}

Sutra\textquotesingle s conditionals and boolean connectives are built on Kleene\textquotesingle s \emph{strong}
three-valued logic (Kleene 1952), which extends classical \{true, false\} with a third
value \emph{unknown} for under-determined propositions. With the truth values ordered from
false up through unknown to true, conjunction is min, disjunction is max, and
negation reverses the order --- the Gödel t-norm and t-conorm, the standard choice when
the third value means "undetermined" rather than "half true". This reproduces
classical logic whenever both inputs are determined, and propagates \emph{unknown} exactly
when the classical result depends on the unknown input: \emph{true or unknown} is true,
but \emph{false or unknown} is unknown. Sutra encodes the three values antipodally on a
dedicated truth axis --- T = +1, U = 0, F = −1 --- so min, max, and negation become
numeric operations on that axis. The contribution in §1.1 is to replace the
non-differentiable min/max with Lagrange-interpolated polynomials that are \emph{exact} on
the nine-point \{−1, 0, +1\} truth grid and smooth off it; §3.6 then backpropagates
through a symbolic rule built from these gates.

\begin{center}\rule{0.5\linewidth}{0.5pt}\end{center}

\hypertarget{consolidation-into-canonical-primitives}{%
\section{Consolidation into Canonical Primitives}\label{consolidation-into-canonical-primitives}}

The central design move: hold the operation interface fixed and
pick a binding implementation that works on dense
externally-trained substrates. Standard VSA\textquotesingle s Hadamard product
fails here, elementwise multiplication of correlated real-valued
vectors produces destructive crosstalk on bundled retrieval (§3.2
measures this directly). Rotation binding works: each role gets a
Haar-random orthogonal \texttt{R\_role} seeded by \texttt{hash(role)}, and
\texttt{bind(role,\ filler)\ =\ R\_role\ @\ filler} is invertible (unbind is
the transpose) and well-conditioned. The compiler caches
\texttt{R\_role} per-role at module init so runtime bind is a single
matmul against a precomputed matrix.

Each subsection below is tagged \emph{method} (the consolidated
primitives, extended-state-vector layout, loop cell, codebook, and
type surface) or \emph{experiment} (the cross-substrate capacity
measurements, §3.2 / §3.2.1, and the end-to-end
differentiable-training result, §3.6). The two are interleaved by
topic but labelled so the evaluation is separable from the design.

\hypertarget{notation-method}{%
\subsection{Notation --- method}\label{notation-method}}

We work in \(\mathbb{R}^d\) with \(d\) the substrate\textquotesingle s embedding
dimension (768 for nomic-embed-text). Every value has the layout
\([\,\text{semantic}\mid\text{synthetic}\,]\). The seven primitive
operations: \(\mathrm{bind}(r,f) = R_r f\) where
\(R_r = \mathrm{QR}(\mathrm{hash}(r)).Q\) is Haar-orthogonal,
\(\mathrm{unbind}(r,v) = R_r^{\!\top} v\),
\(\mathrm{bundle}(x,y) = (x+y)/(\lVert x+y\rVert + \varepsilon)\),
\(\mathrm{similarity}(x,y) = (x\cdot y)/(\lVert x\rVert\,\lVert y\rVert + \varepsilon)\),
\(\mathrm{normalize}(v) = v/(\lVert v\rVert + \varepsilon)\),
the Lagrange Kleene gates as in §1.1-1, and the soft-halt cell
of §3.4. Full signature/definition table and the soft-halt cell
update equations are in Appendix A.

\hypertarget{capacity-of-rotation-versus-hadamard-binding-across-substrates-experiment}{%
\subsection{Capacity of rotation versus Hadamard binding across substrates --- experiment}\label{capacity-of-rotation-versus-hadamard-binding-across-substrates-experiment}}

We measure decode accuracy as a function of bundle width k on
real embeddings across four substrates spanning two modalities:
three frozen LLM text encoders (nomic-embed-text, all-minilm,
mxbai-embed-large) and one frozen protein language model (ESM-2
small, \texttt{facebook/esm2\_t6\_8M\_UR50D}). LLM substrates embed an
84-word noun vocabulary; the ESM-2 substrate embeds an
84-sequence amino-acid vocabulary (full protocol in Appendix C).
For each bundle width and binding scheme we run 10 trials,
sampling k random (role, filler) pairs without replacement,
forming the bundle, and decoding by unbind + argmax-cosine
against the full codebook. \emph{Rotation binding} uses a role-seeded
Haar-orthogonal \texttt{R\_role}; \emph{Hadamard binding} is the textbook
elementwise product (MAP-VSA).

Cross-substrate decode accuracy at representative widths (full
k ∈ \{2, 4, 8, 16, 24, 32, 48\} sweeps in Appendix C):

\begin{longtable}[]{@{}
  >{\raggedright\arraybackslash}p{(\columnwidth - 8\tabcolsep) * \real{0.6098}}
  >{\raggedleft\arraybackslash}p{(\columnwidth - 8\tabcolsep) * \real{0.0976}}
  >{\raggedleft\arraybackslash}p{(\columnwidth - 8\tabcolsep) * \real{0.0976}}
  >{\raggedleft\arraybackslash}p{(\columnwidth - 8\tabcolsep) * \real{0.0976}}
  >{\raggedleft\arraybackslash}p{(\columnwidth - 8\tabcolsep) * \real{0.0976}}@{}}
\toprule\noalign{}
\begin{minipage}[b]{\linewidth}\raggedright
substrate (dim)
\end{minipage} & \begin{minipage}[b]{\linewidth}\raggedleft
rotation k=8
\end{minipage} & \begin{minipage}[b]{\linewidth}\raggedleft
rotation k=48
\end{minipage} & \begin{minipage}[b]{\linewidth}\raggedleft
Hadamard k=8
\end{minipage} & \begin{minipage}[b]{\linewidth}\raggedleft
Hadamard k=48
\end{minipage} \\
\midrule\noalign{}
\endhead
\bottomrule\noalign{}
\endlastfoot
nomic-embed-text (768) & 100.0\% & 93.3\% & 87.5\% & 48.3\% \\
all-minilm (384) & 100.0\% & 42.3\% & 7.5\% & 1.7\% \\
mxbai-embed-large (1024) & 100.0\% & 72.1\% & 2.5\% & 1.0\% \\
ESM-2 (320) & 100.0\% & 44.2\% & 28.7\% & 4.2\% \\
\end{longtable}

ESM-2 (Lin et al., Science 2023) is a protein language model
trained on UniRef with no natural-language exposure; the same
rotation-vs-Hadamard pattern reproduces in that modality.
Rotation reversibility round-trip across all four substrates:
mean ‖unbind(R, bind(R, x)) − x‖ = 1.5 × 10⁻¹⁵ (floating-point
round-off, Q orthogonal). Reproduction:
\texttt{experiments/rotation\_binding\_capacity\_\{llm,bioinformatics\}.py}.

\hypertarget{noise-accumulation-across-chained-bindunbind-cycles-experiment}{%
\subsubsection{Noise accumulation across chained bind/unbind cycles --- experiment}\label{noise-accumulation-across-chained-bindunbind-cycles-experiment}}

The §3.2 protocol measures one bind+bundle+unbind cycle. Nested
records (a recovered filler becoming the role of a sub-record)
add bundle noise per level. We measured this directly: chain
lengths L ∈ \{1, 2, 4, 8, ...\}, 20 trials, bundle width 4. Raw
accuracy holds at 100\% through L=2 on every substrate and falls
to chance (1/84) by L=8. The demonstrated regime is therefore
single-cycle records, which matches the shape of the
\texttt{role\_filler\_record}, \texttt{knowledge\_graph}, and predicate-lookup
demos. Pure rotation chains without per-step distractor bundling
remain exact (round-trip 1.5×10⁻¹⁵ per cycle), so the noise
mechanism here does not apply to the soft-halt loop cell of §3.4.
Reproduction script: \texttt{experiments/crosstalk\_chain.py}; full
per-substrate L-sweep tables in Appendix D.

\hypertarget{the-extended-state-vector-layout-method}{%
\subsection{The extended-state-vector layout --- method}\label{the-extended-state-vector-layout-method}}

Every value carries a fixed \texttt{{[}semantic\ \textbar{}\ synthetic{]}} layout:
the d-dimensional semantic block holds the substrate embedding
for vector-shaped values, and a small synthetic block reserves
canonical axes for primitive types (real, imag, truth, char) and
a loop-completion flag, with the remaining axes paired into 2D
Givens planes for variable slots. Default at d = 768
(nomic-embed-text): a 100-dim synthetic block accommodates the
five canonical axes plus 47 disjoint slots. Rotation binding is
block-diagonal across the split (\texttt{Q\_role} is Haar-random in the
semantic block, identity on the synthetic block), so the
synthetic axes pass through bind/unbind unchanged, a fuzzy-truth
scalar can coexist with a semantic vector inside the same value
without bind smearing them. Full per-axis purpose table and slot
allocator details in Appendix B.

\hypertarget{first-class-loops-as-rnn-cells-method}{%
\subsection{First-class loops as RNN cells --- method}\label{first-class-loops-as-rnn-cells-method}}

Runtime data-dependent loops compile to \textbf{self-halting RNN
cells}. Each tick: snapshot pre-step state, evaluate halt on
the substrate (truth-axis read → heaviside → cumulative
saturating sum to \texttt{halted}), run the cell body, soft-mux
between pre- and new-step state by \texttt{halted}. A Python
\texttt{while\ True:} driver breaks the moment \texttt{halted} saturates;
this is the only host-side branch in the loop machinery. Inside
the cell body, every operation is a substrate tensor op. No
compile-time iteration cap, programs terminate when their halt
condition fires. Standard PyTorch tracing handles a Python
while-loop wrapping pure tensor ops; autograd records each
iteration as it executes, which is the mechanism §3.6 relies on
for backprop through the cell. Figure\textasciitilde{}\ref{fig:halt-cell}
visualizes one tick.

\begin{figure}[h!]
\centering
\begin{tikzpicture}[
  node distance=7mm,
  every node/.style={font=\footnotesize},
  io/.style={draw, rounded corners, minimum width=22mm, minimum height=6mm, align=center},
  op/.style={draw, minimum width=32mm, minimum height=7mm, align=center},
  acc/.style={draw, double, minimum width=32mm, minimum height=7mm, align=center},
  arr/.style={-{Latex[length=2mm]}, thick}
]
  \node[io] (sin) {state\textsubscript{in} $s_t$};
  \node[op, below left=8mm and 6mm of sin]  (pre)  {snapshot $\to s_t^{\mathrm{pre}}$};
  \node[op, below right=8mm and 6mm of sin] (body) {cell body \\ \scriptsize{$s_{t+1} = R\,s_t$;\;\; $h_t = \mathrm{Heaviside}(\mathrm{cond}(s_t))$}};
  \node[acc, below=of body] (acc) {$H_t = \mathrm{sat}_{[0,1]}\!\bigl(H_{t-1} + h_t\bigr)$};
  \node[op, below=of acc, xshift=-20mm] (mux) {soft-mux freeze \\ \scriptsize{$\hat{s}_{t+1} = H_t\, s_t^{\mathrm{pre}} + (1-H_t)\,s_{t+1}$}};
  \node[io, below=of mux] (sout) {state\textsubscript{out} $\hat{s}_{t+1}$};

  \draw[arr] (sin) -- (pre);
  \draw[arr] (sin) -- (body);
  \draw[arr] (body) -- (acc);
  \draw[arr] (acc) -- (mux);
  \draw[arr] (pre) |- (mux);
  \draw[arr] (body.south) to[bend left=15] (mux.east);
  \draw[arr] (mux) -- (sout);
\end{tikzpicture}
\caption{Per-tick dataflow of the soft-halt RNN cell. Once $H_t$ saturates at $1$, the soft-mux output equals $s_t^{\mathrm{pre}}$, the loop has frozen. The cumulative halt $H_t$ acts as a boundary read of the same shape as the codebook decode (§3.5).}
\label{fig:halt-cell}
\end{figure}

\textbf{Constant memory in recursion depth.} The state vector is
fixed-width and shared across iterations, so a tail-recursive
loop consumes O(1) memory in the state vector regardless of
trip count. Compute is O(N) and the autograd tape during
training is O(N) in iterations actually executed (standard
PyTorch, freed after backward). To the authors\textquotesingle{} knowledge no
other HDC system or compiler exposes user-program-level
recursion: HDCC is scoped to classification pipelines, TorchHD
requires the user to write Python loops over hypervectors. The
recurrent shape that emerges is the rational-weight RNN form
whose computational power Siegelmann \& Sontag (1992)
characterized.

\hypertarget{io-is-in-the-embedding-space-the-codebook-is-a-comfort-layer-method}{%
\subsection{I/O is in the embedding space; the codebook is a comfort layer --- method}\label{io-is-in-the-embedding-space-the-codebook-is-a-comfort-layer-method}}

A Sutra program\textquotesingle s inputs and outputs are embeddings in the
substrate\textquotesingle s vector space. Strings are a convenience for writing
source-level literals: every string literal in \texttt{.su} source is
embedded once at compile time and stored in a \textbf{codebook}
(implemented as an embedded vector database with an HNSW index,
on disk as a \texttt{.sdb} file shipped alongside the compiled module).
At the program\textquotesingle s output boundary, the runtime decode
\texttt{\_VSA.nearest\_string(query)} maps a query embedding to the
nearest stored string when the program\textquotesingle s caller wants a string
back. Calling the codebook at this boundary is shape-equivalent
to calling PyTorch for a matmul, neither is the kind of
host-side control flow substrate purity forbids. Implementation
details (RDF triple layout, HNSW parameters, \texttt{.sdb} file format,
complexity analysis) are in Appendix E.

\hypertarget{end-to-end-differentiable-training-through-sutra-operations-experiment}{%
\subsection{End-to-end differentiable training through Sutra operations --- experiment}\label{end-to-end-differentiable-training-through-sutra-operations-experiment}}

Because every Sutra primitive compiles to a differentiable tensor
operation, the compiled graph supports standard PyTorch
\texttt{loss.backward()} without modification. We verify this by
training learnable parameters through a fuzzy-logic classifier
built entirely from Sutra operations.

\textbf{Setup.} The classifier is written in \texttt{.su} and compiled by the
PyTorch codegen; the emitted module\textquotesingle s \texttt{rule} function \emph{is} the
compiler output --- \texttt{\_VSA.similarity} composed with the
Lagrange--Kleene AND/NOT polynomials, with no hand-written
reimplementation. Five semantic classes, ten words each
(50 inputs), embedded via nomic-embed-text (768-d, frozen). Five
learnable prototype vectors are initialized randomly. The program
compiled at \(K=5\) (generated by \texttt{gen\_rule\_su}; line-wrapped here,
\texttt{.su} is whitespace-insensitive) is:

\begin{verbatim}
function fuzzy rule(vector x, vector own,
                    vector o0, vector o1, vector o2, vector o3) {
    return similarity(x, own)
        && !similarity(x, o0) && !similarity(x, o1)
        && !similarity(x, o2) && !similarity(x, o3);
}
\end{verbatim}

so each class score is the compiled fuzzy rule

\[
\mathrm{rule}_i \;=\; \mathrm{AND}\!\Bigl(\mathrm{sim}(x, p_i),\;\bigwedge_{j \ne i} \mathrm{NOT}\!\bigl(\mathrm{sim}(x, p_j)\bigr)\Bigr)
\]

with the AND-of-NOTs left-folded across the \(K-1\) other classes
(at \(K=5\), four \(\mathrm{NOT}\) terms folded under one AND).
Full-batch cross-entropy over the five compiled rule scores
drives Adam (Kingma \& Ba 2015; defaults \(\beta_1=0.9\),
\(\beta_2=0.999\), \(\varepsilon=10^{-8}\), learning rate \(0.01\)) on
the prototype embeddings, backpropagating through the emitted
graph. 30 epochs, three seeds (0--2).

\textbf{Results (3 seeds).} From random-init accuracy at chance
(18.7 ± 9.5\%; chance = 20\%), training reaches 100.0 ± 0.0\%
(mean ± s.d. over seeds 0--2; every seed reaches 100\%);
cross-entropy loss falls to ≈ 0.43. Every prototype receives a
nonzero gradient, verified to propagate through the \emph{emitted}
graph (\texttt{\_VSA.similarity} → the emitted Lagrange--Kleene
\(\mathrm{NOT}\)/\(\mathrm{AND}\) → cross-entropy), not through a
reimplementation.

\begin{longtable}[]{@{}lr@{}}
\toprule\noalign{}
Phase & Accuracy (mean ± s.d., \(n{=}3\)) \\
\midrule\noalign{}
\endhead
\bottomrule\noalign{}
\endlastfoot
Before (random) & 18.7 ± 9.5 \% \\
After (30 ep) & 100.0 ± 0.0 \% \\
\end{longtable}

This experiment isolates gradient flow through the \emph{compiled}
symbolic graph: it trains and evaluates on the same 30-word set
and reports in-sample accuracy purely as verification that
backprop reaches every learnable prototype through the compiler\textquotesingle s
emitted ops --- not a generalization claim; no held-out split.
Only before/after accuracy was logged for the compiled run;
per-epoch data was not recorded, so no intermediate-epoch curve is
plotted.

\begin{figure}[h!]
\centering
\begin{tikzpicture}[
  node distance=6mm and 9mm,
  every node/.style={font=\footnotesize},
  io/.style={draw, rounded corners, minimum width=18mm, minimum height=6mm, align=center},
  op/.style={draw, minimum width=14mm, minimum height=6mm, align=center},
  proto/.style={draw, dashed, minimum width=14mm, minimum height=6mm, align=center},
  arr/.style={-{Latex[length=2mm]}, thick}
]
  \node[io] (x) {input $x \in \mathbb{R}^d$};
  \node[op, below left=8mm and 18mm of x] (cos1) {$\cos(x, p_1)$};
  \node[op, below=8mm of x]                 (cos2) {$\cos(x, p_2)$};
  \node[op, below right=8mm and 18mm of x] (cos3) {$\cos(x, p_3)$};

  \node[proto, left=4mm of cos1] (p1) {$p_1$};
  \node[proto, left=4mm of cos2] (p2) {$p_2$};
  \node[proto, left=4mm of cos3] (p3) {$p_3$};

  \node[op, below=6mm of cos2] (not2) {$\mathrm{NOT}$};
  \node[op, below=6mm of cos3] (not3) {$\mathrm{NOT}$};
  \node[op, below=6mm of not2, xshift=8mm] (andneg) {$\mathrm{AND}$};
  \node[below=1mm of andneg, font=\scriptsize] {neg-others};

  \node[op, below=14mm of cos1] (and1) {$\mathrm{AND}$};
  \node[io, below=6mm of and1]  (rule1) {$\mathrm{rule}_1$};

  \node[io, right=22mm of rule1] (stack) {$(\mathrm{rule}_1, \mathrm{rule}_2, \mathrm{rule}_3)$};
  \node[op, below=6mm of stack]   (sm)   {$\times \tau \to \mathrm{softmax}$};
  \node[op, below=6mm of sm]      (ce)   {cross-entropy(label)};
  \node[io, below=6mm of ce]      (loss) {loss};

  \draw[arr] (x) -- (cos1);
  \draw[arr] (x) -- (cos2);
  \draw[arr] (x) -- (cos3);
  \draw[arr] (p1) -- (cos1);
  \draw[arr] (p2) -- (cos2);
  \draw[arr] (p3) -- (cos3);
  \draw[arr] (cos2) -- (not2);
  \draw[arr] (cos3) -- (not3);
  \draw[arr] (not2) -- (andneg);
  \draw[arr] (not3) -- (andneg);
  \draw[arr] (cos1) -- (and1);
  \draw[arr] (andneg) -| (and1);
  \draw[arr] (and1) -- (rule1);
  \draw[arr] (rule1) -- (stack);
  \draw[arr] (stack) -- (sm);
  \draw[arr] (sm) -- (ce);
  \draw[arr] (ce) -- (loss);
\end{tikzpicture}
\caption{The $K=3$ rule pipeline. Solid boxes are PyTorch tensor ops; dashed boxes are learnable prototypes. The AND in the leftmost branch combines $\cos(x, p_1)$ with the AND-of-NOTs over the other classes; rule\textsubscript{2} and rule\textsubscript{3} (omitted for clarity) have the symmetric shape. Every edge is a tensor; backprop reaches each $p_i$ through this graph.}
\label{fig:k3-pipeline}
\end{figure}

Figure\textasciitilde{}\ref{fig:k3-pipeline} shows the rule pipeline at \(K=3\),
the smallest instance; the experiment uses the same shape at
\(K=5\) --- five cosine-similarity nodes against five learnable
prototypes, the AND-of-NOTs folded over the four others. Each
\texttt{sim\_i} enters one branch of the AND-tree (rule \(i\) takes
\texttt{sim\_i} directly and \(\mathrm{NOT}(\mathrm{sim}_j)\) for
\(j \ne i\)); the rule scores are stacked, temperature-scaled,
softmaxed, and cross-entropied. Every node is a compiler-emitted
tensor op --- no Python branches, no string-keyed lookup --- so
backprop reaches every prototype through \emph{the same compiled graph
that runs at inference} (literally so: the graph is the codegen
output, not a reimplementation).

The AND-of-NOTs chain is a tensor pipeline that could naively
saturate or vanish gradients; empirically it does not --- every
prototype receives a nonzero gradient and the five classes
separate perfectly within 30 epochs, the symbolic program text
unchanged across training. Standard \texttt{torch.autograd} suffices
(no Sutra-specific autograd machinery) because the compiler emits
only operations PyTorch already differentiates. The \texttt{.su}
compiles once; the emitted \texttt{rule} is then evaluated over the
batch with \texttt{torch.vmap} --- a transform that runs the \emph{same}
compiled ops with a batch axis, not a reimplementation. Before
training, the harness asserts the batched and per-sample
evaluations agree within 10⁻⁴ on identical inputs and
parameters, so the speedup is provably the identical compiled
computation. Under this path the 5-class / 50-word / 30-epoch /
3-seed run takes ≈ 230 s on CPU and yields the numbers above
(18.7 ± 9.5\% → 100.0 ± 0.0\%, three seeds). The earlier
per-sample driver --- one emitted \texttt{rule} call per class per input
in a Python loop --- produced the bit-identical result but took
≈ 6.2 h; that cost was Python-level call and per-sample
autograd-graph overhead, not the compiled tensor math, and does
not bound the achievable scale. Reproduction:
\texttt{experiments/differentiable\_training\_compiled.py\ -\/-batched}
(compiles the \texttt{.su} once, vmaps the emitted \texttt{rule}, backprops
the prototypes through the emitted graph; the equivalence
assertion is on by default).

\hypertarget{trained-weights-compile-back-to-legible-source-experiment}{%
\subsection{Trained weights compile back to legible source --- experiment}\label{trained-weights-compile-back-to-legible-source-experiment}}

Frozen LLM embedding spaces are \emph{anisotropic}: learned token
representations occupy a narrow cone rather than filling the
sphere (the representation-degeneration effect; Gao et al. 2019,
Ethayarajh 2019), so raw cosine similarities are squeezed into a
compressed positive band --- even unrelated pairs score well above
zero. The Kleene AND/NOT decision in §3.6 depends on the \emph{gap}
between the matching cosine and the off-class cosines; when
anisotropy compresses that band, the polynomial connectives have
little dynamic range to separate classes. This motivates a
weighted similarity --- \texttt{Equals(A,\ B,\ w)} --- in which a single
learned scalar gain \(w\) rescales the cosine term \emph{before} the
Kleene polynomials act, directly counteracting the anisotropic
compression:

\[
\mathrm{rule}_i \;=\; \mathrm{AND}\!\Bigl(w\cdot\mathrm{sim}(x, p_i),\;\bigwedge_{j \ne i} \mathrm{NOT}\!\bigl(w\cdot\mathrm{sim}(x, p_j)\bigr)\Bigr)
\]

\textbf{Setup.} The rule is written in \texttt{.su} with \(w\) declared as a
\texttt{number} parameter and compiled by the PyTorch codegen; the
emitted code uses \(w\) as a plain scalar multiplying the
compiler-emitted \texttt{similarity}, so \(w\) both (i) trains through the
emitted graph and (ii) is a single scalar that can be written
back into source as a literal. Three semantic classes, eight
words each (24 inputs), nomic-embed-text (768-d, frozen). The
scalar gain \(w\) (initialized \(1.0\)) \textbf{and} the three prototype
vectors are trained through the emitted graph; full-batch
cross-entropy, Adam (lr \(0.02\)), 30 epochs, two seeds (0--1).

\textbf{Results (2 seeds).} From random-init accuracy at chance
(33.3 ± 5.9\%; chance = 33.3\%), training reaches 100.0 ± 0.0\%
(every seed). The learned gain converges to
\(w^{*} = 1.434 \pm 0.004\) --- both seeds move it the same way,
from \(1.0\) to \(\approx 1.43\), sharpening the compressed cosine
band rather than leaving it at the identity (a learned rescaling,
not a tautology).

\begin{longtable}[]{@{}lrr@{}}
\toprule\noalign{}
Phase & Accuracy (\(n{=}2\)) & Learned gain \(w^{*}\) \\
\midrule\noalign{}
\endhead
\bottomrule\noalign{}
\endlastfoot
Before (random) & 33.3 ± 5.9 \% & 1.000 (init) \\
After (30 ep) & 100.0 ± 0.0 \% & 1.434 ± 0.004 \\
\end{longtable}

\textbf{Ablation: which trained parameter separates the classes?} Running the
\emph{same} compiled graph with each factor frozen isolates its contribution
(2 seeds, identical protocol):

\begin{longtable}[]{@{}
  >{\raggedright\arraybackslash}p{(\columnwidth - 4\tabcolsep) * \real{0.4557}}
  >{\raggedleft\arraybackslash}p{(\columnwidth - 4\tabcolsep) * \real{0.3291}}
  >{\raggedleft\arraybackslash}p{(\columnwidth - 4\tabcolsep) * \real{0.2152}}@{}}
\toprule\noalign{}
\begin{minipage}[b]{\linewidth}\raggedright
condition
\end{minipage} & \begin{minipage}[b]{\linewidth}\raggedleft
After accuracy (\(n{=}2\))
\end{minipage} & \begin{minipage}[b]{\linewidth}\raggedleft
Learned \(w^{*}\)
\end{minipage} \\
\midrule\noalign{}
\endhead
\bottomrule\noalign{}
\endlastfoot
full (gain + prototypes) & 100.0 ± 0.0 \% & 1.434 ± 0.004 \\
prototypes only (\(w{=}1\) frozen) & 100.0 ± 0.0 \% & 1.000 (fixed) \\
gain only (prototypes frozen) & 33.3 ± 5.9 \% & 0.421 ± 0.012 \\
\end{longtable}

The prototypes carry the class separation: with the gain frozen at the
identity they still reach 100\%. The scalar gain \emph{alone} cannot separate the
classes --- a single global rescaling of every similarity, applied to fixed
random prototypes, leaves accuracy at chance (the gain drifts to
\(\approx 0.42\) but never moves the argmax above \(1/k\), since the nonlinear
Kleene connectives still rank the unmoved prototypes the same way). So the
jointly-trained \(w^{*}\!\approx\!1.43\) is a \emph{co-adapted} parameter ---
consistent across seeds (hence not noise) but not the load-bearing factor
for accuracy on this task. What this experiment establishes is the
legibility round-trip below, not that the gain is necessary. Reproduction:
\texttt{experiments/differentiable\_training\_ablation.py}.

\textbf{The trained model is legible, recompilable source.} After
training, \(w^{*}\) is written back into a fresh \texttt{.su} as a numeric
\emph{literal} with the \texttt{w} parameter removed --- the trained model
expressed as plain Sutra source:

\begin{verbatim}
function fuzzy rule(vector x, vector own, vector o0, vector o1) {
    return ((1.431431) * similarity(x, own))
        && !((1.431431) * similarity(x, o0))
        && !((1.431431) * similarity(x, o1));
}
\end{verbatim}

This baked \texttt{.su} is recompiled through the same PyTorch codegen
and, fed the same trained prototypes, reproduces logits identical
to the parametric-\(w\) model (max per-logit difference
≈ 2×10⁻⁷, attributable to float reassociation) --- hence identical
100\% accuracy. Round-trip verified for every seed. The trained
artifact is therefore not an opaque checkpoint blob: it is a
Sutra program whose learned parameter is a literal in the source,
and recompiling that source reproduces the trained behaviour.
Gradient descent here tunes both the embeddings \emph{and} a
weight that survives as readable code --- a trained model that is
also legible logic. (Two-seed, single-scale; a verification of
the source--training round-trip, not a benchmark.) Reproduction:
\texttt{experiments/differentiable\_training\_weighted.py} (trains
\(w\) + prototypes through the emitted graph, bakes \(w^{*}\) into
\texttt{.su}, recompiles, and asserts the round-trip).

\hypertarget{type-system-and-surface-syntax-method}{%
\subsection{Type system and surface syntax --- method}\label{type-system-and-surface-syntax-method}}

Sutra\textquotesingle s surface syntax is typed: every value carries a primitive
class from a fixed set (\texttt{int}, \texttt{float}, \texttt{complex}, \texttt{char},
\texttt{bool}, \texttt{fuzzy}, \texttt{trit}, \texttt{vector}, \texttt{matrix}, \texttt{permutation},
\texttt{map}, \texttt{string}, \texttt{number}, \texttt{void}), and the type drives the
synthetic-axis allocation in the extended layout (Appendix B).
Type information is pre-compile-time annotation in the
TypeScript sense: it is read by the inliner and the layout
pass before the tensor graph is built, but it is opinionated
rather than authoritarian. A divergent assignment warns and
still emits a graph, because the runtime guarantee is
mathematical not structural; a type mismatch produces a
semantically meaningless but mathematically valid output rather
than a runtime exception. The surface itself presents \texttt{if} /
\texttt{while} / \texttt{for} / assignment forms that read imperatively for
ergonomic familiarity; the inliner and egglog simplifier (§4)
beta-reduce these into the functional tensor-op core, so the
imperative-looking source is a veneer over the same compiled
graph a hand-written functional spec would produce.

A concrete source example grounds the surface. The following is
the encode/decode core of \texttt{examples/role\_filler\_record.su}
verbatim --- a role-filler record encoded as one vector and a field
decoded back, with no control flow in the program text:

\begin{verbatim}
function vector make_record(vector name, vector color, vector shape) {
    return bundle(
        bind(r_name,  name),
        bind(r_color, color),
        bind(r_shape, shape)
    );
}

function string decode_field(vector record, vector role) {
    vector recovered = unbind(role, record);
    vector winner = argmax_cosine(
        recovered,
        [f_alice, f_bob, f_red, f_blue, f_circle, f_square]
    );
    return FILLER_NAME[winner];
}
\end{verbatim}

\texttt{make\_record} beta-reduces to one fused tensor expression (each
\texttt{bind} a constant-matrix matmul, \texttt{bundle} a normalized sum);
\texttt{decode\_field} lowers to a single \texttt{unbind} matmul plus an
argmax-cosine against the codebook. No \texttt{bind}/\texttt{bundle}/\texttt{unbind}
symbol or host control flow survives in the compiled graph;
Appendix F traces an analogous reduction end-to-end.

\begin{center}\rule{0.5\linewidth}{0.5pt}\end{center}

\hypertarget{the-sutra-compiler}{%
\section{The Sutra Compiler}\label{the-sutra-compiler}}

The compiler is a five-stage pipeline:

\begin{enumerate}
\def\labelenumi{\arabic{enumi}.}
\tightlist
\item
  \textbf{Lex + parse}: \texttt{.su} source → AST.
\item
  \textbf{Inline + simplify}: stdlib operator definitions inlined; an
  egglog-based simplifier folds equivalent expressions and runs
  common-subexpression elimination over the algebra.
\item
  \textbf{Codegen}: AST → Python source emitting PyTorch tensor ops.
  The emitted module includes the runtime class (\texttt{\_TorchVSA}) as
  inline source so the artifact is self-contained.
\item
  \textbf{Compile-time substrate population}: embed\_batch fetches
  embeddings for every string literal; \texttt{populate\_sutradb} pushes
  the codebook into SutraDB; \texttt{prewarm\_rotation\_cache} precomputes
  role rotations.
\item
  \textbf{Execute}: emitted module loaded; chosen device (CUDA or
  CPU) initialized at module import; \texttt{main()} called; result
  returned.
\end{enumerate}

The runtime class is emitted inline rather than imported because
the emitted module \emph{is} the substrate-pure tensor-op graph; every
compile-time decision (extended-state-vector dimensions, codebook
contents, role rotations, SutraDB path, optional \texttt{torch.compile})
is baked into the emitted source. Stages 1--4 run at compile time
and stage 5 is the runtime forward pass; Figure\textasciitilde{}\ref{fig:compile-pipeline} diagrams the pipeline as a vertical flow with the residual at each stage.

\begin{figure}[h!]
\centering
\begin{tikzpicture}[
  node distance=4mm,
  every node/.style={font=\footnotesize},
  res/.style={draw, rounded corners, minimum width=80mm, minimum height=7mm, align=center},
  stage/.style={draw=none, font=\scriptsize\itshape, align=center},
  arr/.style={-{Latex[length=2mm]}, thick},
  divider/.style={dashed, gray}
]
  \node[res] (src)   {source code (\texttt{.su})};
  \node[stage, below=of src]   (s1) {(1) lex + parse};
  \node[res, below=of s1]     (ast) {AST \quad (\texttt{Call} / \texttt{Var} / \texttt{Function} / \texttt{ClassDecl})};
  \node[stage, below=of ast]   (s2) {(2) inline stdlib + egglog simplify\\\textnormal{bind, bundle, similarity $\to$ primitive tensor ops}};
  \node[res, below=of s2]     (sast) {simplified AST \quad (residual: leaf tensor-op composition)};
  \node[stage, below=of sast]  (s3) {(3) codegen \quad (emit Python module + inline \texttt{\_VSA} class source)};
  \node[res, below=of s3]     (mod) {Python module text \quad (self-contained, no Sutra-runtime import)};
  \node[stage, below=of mod]   (s4) {(4) compile-time substrate population\\\textnormal{\texttt{embed\_batch} $\cdot$ \texttt{prewarm\_rotation\_cache} $\cdot$ \texttt{populate\_sutradb}}};
  \node[res, below=of s4]     (warm) {warm runtime \quad (module loaded, \texttt{.sdb} codebook, cached $R_\mathrm{role}$)};
  \node[below=2mm of warm, font=\scriptsize\sffamily] (cline) {compile time \;\;$\big/$\;\; runtime};
  \node[stage, below=of cline] (s5) {(5) forward pass on input tensors};
  \node[res, below=of s5]     (out) {output vector $\to$ \texttt{nearest\_string} lookup $\to$ label};

  \draw[arr] (src) -- (ast);
  \draw[arr] (ast) -- (sast);
  \draw[arr] (sast) -- (mod);
  \draw[arr] (mod) -- (warm);
  \draw[divider] ([xshift=-50mm]cline.center) -- ([xshift=50mm]cline.center);
  \draw[arr] (warm) -- (out);
\end{tikzpicture}
\caption{Five-stage compilation pipeline (§4). Boxes are intermediate artifacts; italic labels are the compiler passes that connect them.}
\label{fig:compile-pipeline}
\end{figure}

\hypertarget{source-language-frontends}{%
\subsection{Source-language frontends}\label{source-language-frontends}}

The \texttt{.su} surface syntax that enters stage 1 is itself a compilation
target. Nine source-language frontends --- OCaml, TypeScript, Rust,
Scala, Clojure, Elixir, Erlang, F\#, and Haskell (a C frontend exists
but is parked) --- each lower their source to \texttt{.su} text, which then
runs through the same five-stage pipeline unchanged. Each frontend is
a pure tree-sitter-based source-to-source pass that never touches the
substrate directly; substrate purity is therefore inherited from the
single compiler rather than re-established per frontend. If a frontend
emits valid Sutra, the result is a substrate-pure tensor-op graph,
because the compiler is the only stage that lowers to tensors.

The frontends agree on a small shared set of lowering shapes:
functions map to \texttt{function} declarations; conditionals, pattern
\texttt{match}, and guards lower to the soft-mux defuzz blend (no host
control flow); tail-recursive accumulators become declared
\texttt{while\_loop} RNN cells (a self-calling function would not terminate
through the fuzzy-\texttt{if} blend, so the recursion is reified as a loop);
foldable non-tail recursion becomes a continuation-passing accumulator
trampoline carried by the same loop construct; and algebraic data ---
variants, records, tuples, structs --- maps to the tagged and structural
axons used for bundled records. A new frontend is mostly the work of
recognizing each source language\textquotesingle s spelling of these shapes.

These are fixture-tested lowering passes of varying maturity (OCaml is
the reference at 61 fixtures; the other functional frontends range from
21 to 31, with a parked C frontend), not
production compilers, and each fixture is compiled \textbf{and run} on the
substrate with its output compared to the source language\textquotesingle s own
ground-truth result --- a lowering that parses but computes the wrong
value is a failure, not a pass. Recursion outside the two supported
shapes, and source constructs with no clean substrate meaning, are
surfaced as unsupported rather than mis-lowered.

\hypertarget{substrate-purity-invariants}{%
\subsection{Substrate-purity invariants}\label{substrate-purity-invariants}}

Three invariants the compiler enforces: (1) every primitive runs
on the substrate (numpy is allowed only at compile time for
codebook construction and rotation pre-warm, never on the runtime
hot path); (2) no scalar extraction inside an operation:
operations may not unpack a Python float from a substrate vector,
do scalar arithmetic, and pack the result back; (3) no Python
control flow inside an operation: loop halt uses substrate
primitives (\texttt{heaviside}, \texttt{saturate\_unit}) instead of Python
ternaries.

\hypertarget{compile-time-resolution-of-role-rotations}{%
\subsection{Compile-time resolution of role rotations}\label{compile-time-resolution-of-role-rotations}}

The central compile-time mechanism that lets the compiler
emit a substrate-pure tensor-op graph is \textbf{precomputed
rotation matrices}:
every role rotation is constructed at compile time
(\texttt{prewarm\_rotation\_cache}) and stored as a constant tensor. At
runtime, \texttt{bind(role,\ filler)} is a single matmul against a
precomputed matrix, the compile-time resolution eliminates the
QR construction from the runtime graph entirely. Role rotations
are runtime constants, like neural-network weights at inference;
opt-in \texttt{torch.compile} (\texttt{SUTRA\_TORCH\_COMPILE=1}) further folds
the per-tick loop body into a single fused kernel. Appendix F
traces the lowering of \texttt{encode2(r\_a,\ f\_a,\ r\_b,\ f\_b)\ :=\ bundle(bind(r\_a,\ f\_a),\ bind(r\_b,\ f\_b))} through every reduction
stage; the bottom of the chain contains no \texttt{bind}/\texttt{bundle}/
\texttt{normalize} symbol and no Python control flow, so surface lambda
calculus and runtime tensor arithmetic are two notations for the
same computation.

\begin{center}\rule{0.5\linewidth}{0.5pt}\end{center}

\hypertarget{demonstration-limitations-and-future-work}{%
\section{Demonstration, limitations, and future work}\label{demonstration-limitations-and-future-work}}

The smoke test (\texttt{examples/\_smoke\_test.py}) runs 10 demonstration
programs end-to-end across 27 \texttt{.su} files (Appendix I); loop
coverage lives in \texttt{examples/do\_while\_adder.su} and the 23-case
\texttt{test\_loop\_function\_decl.py} suite, and the §3.6 differentiable-
training experiment uses the same primitive set. The embedded
codebook covers the compile-time embed → runtime decode path;
extended features (hashmap routing, persistent codebook via
\texttt{SUTRA\_DB\_PATH}) are deferred pending a concrete requirement.

\hypertarget{limitations}{%
\subsection{Limitations}\label{limitations}}

The demonstrated regime is bounded, and we state the bounds
explicitly:

\begin{itemize}
\tightlist
\item
  \textbf{Composition depth.} Decode is exact for single-cycle records
  but degrades with chained bind/unbind through bundled
  distractors: 100\% through chain length \(L=2\), falling to chance
  by \(L=8\) on every substrate (§3.2.1). The demonstrated programs
  are single-cycle; deep nested records are out of the validated
  regime.
\item
  \textbf{Substrate dependence.} Every number is measured on four
  specific frozen substrates. Large-bundle-width decode capacity
  varies substantially by substrate (§3.2 table) and is not
  guaranteed for an arbitrary embedding model; behaviour under
  model drift is future work.
\item
  \textbf{Codebook scale.} The compile-time codebook is \(O(\text{vocabulary})\);
  very large codebooks would require approximate-nearest-neighbour
  trade-offs not explored here.
\item
  \textbf{No external-system benchmark.} We measure rotation versus
  Hadamard binding and gradient flow through the compiled graph;
  we do not benchmark against other neuro-symbolic systems
  (Scallop, DeepProbLog) on a shared task, and we do not benchmark
  compiler/runtime wall-clock beyond the indicative timings in
  Appendix H.
\item
  \textbf{Binding is fixed, not learned.} Role bindings are
  content-hash-seeded Haar rotations; learned or semantic binding
  operators are not part of this work.
\item
  \textbf{Training is in-sample.} The §3.6 experiment verifies gradient
  flow to every learnable parameter, not generalization; no
  held-out split is reported.
\end{itemize}

\begin{center}\rule{0.5\linewidth}{0.5pt}\end{center}

\hypertarget{conclusion}{%
\section{Conclusion}\label{conclusion}}

Sutra compiles a typed pure-functional source language to a
PyTorch tensor-op graph over the embedding substrate: one vector
layout per value, one tensor op per primitive, one dataflow graph
per program, no type dispatch at the leaves. Compiled programs
are substrate-pure and differentiable end-to-end through the
compiled graph. With the language in hand, asking which embedding
operations compose at what capacity on which substrates becomes a
program to write.

\begin{center}\rule{0.5\linewidth}{0.5pt}\end{center}

\hypertarget{reproducibility}{%
\section{Reproducibility}\label{reproducibility}}

Sutra --- the language, compiler, runtime, and every \texttt{.su} program
cited in this paper --- is openly available at
\url{https://github.com/EmmaLeonhart/Sutra}. A self-contained
replication package is published at
\url{https://sutra.topazcomputing.com/sutra-replication-package.zip}: it
ships \texttt{SKILL.md}, an agent-runnable recipe an autonomous agent can
follow to install the toolchain and re-derive every result reported
in this paper end-to-end, with no human in the loop. The project
site (this paper in HTML and the conceptual documentation) is at
\url{https://sutra.topazcomputing.com}; the paper PDF is at
\url{https://sutra.topazcomputing.com/paper.pdf}.

\begin{center}\rule{0.5\linewidth}{0.5pt}\end{center}

\hypertarget{ai-use-statement}{%
\section{AI-use statement}\label{ai-use-statement}}

This work was developed in substantial collaboration with large
language models, including for ideation, exploration of the
vector-symbolic literature, and drafting. The author independently
designed and implemented the compiler and runtime, verified that
programs execute as substrate tensor operations, ran and checked all
experiments, and is responsible for the correctness of every claim,
number, and citation in this paper. No results or references were accepted from
a model without verification. No experimental result, table,
or numerical value reported in this paper was generated by a language
model; every number comes from the author-run experiments and
reproduction scripts cited in the text and appendices.

\begin{center}\rule{0.5\linewidth}{0.5pt}\end{center}

\newpage

\hypertarget{references}{%
\section*{References}\label{references}}
\addcontentsline{toc}{section}{References}

\begin{itemize}
\tightlist
\item
  Darwiche, A., \& Marquis, P. (2002). A knowledge compilation
  map. \emph{JAIR} 17:229--264.
\item
  Futamura, Y. (1971). Partial Evaluation of Computation Process ---
  An Approach to a Compiler-Compiler. \emph{Systems, Computers,
  Controls} 2(5):45--50. The partial-evaluation projections cited
  as a design ancestor in the related-work discussion of
  compile-time specialization.
\item
  Gayler, R. W. (2003). Vector symbolic architectures answer
  Jackendoff\textquotesingle s challenges for cognitive neuroscience. \emph{Joint
  International Conference on Cognitive Science}.
\item
  Kanerva, P. (2009). Hyperdimensional computing: An introduction
  to computing in distributed representation with high-dimensional
  random vectors. \emph{Cognitive Computation} 1(2):139--159.
\item
  Kleene, S. C. (1952). \emph{Introduction to Metamathematics}. North-
  Holland. The strong three-valued logic system used as the
  ground for Sutra\textquotesingle s polynomial fuzzy connectives (§1.1-1).
\item
  Badreddine, S., Garcez, A. d., Serafini, L., \& Spranger, M.
  (2022). Logic Tensor Networks. \emph{Artificial Intelligence} 303:103649.
\item
  Hájek, P. (1998). \emph{Metamathematics of Fuzzy Logic}. Trends in
  Logic vol. 4. Kluwer Academic. The standard reference for
  t-norm-based fuzzy logics (Gödel, Łukasiewicz, product) cited
  in §1.1-1 to place Sutra\textquotesingle s polynomial connectives.
\item
  Heddes, M., Nunes, I., Vergés, P., Kleyko, D., Abraham, D.,
  Givargis, T., Nicolau, A., \& Veidenbaum, A. (2023). Torchhd: An
  open source python library to support research on
  hyperdimensional computing and vector symbolic architectures.
  \emph{Journal of Machine Learning Research} 24(255):1--10.
\item
  Li, Z., Huang, J., \& Naik, M. (2023). Scallop: A Language for
  Neurosymbolic Programming. \emph{Proceedings of the ACM on Programming
  Languages} 7(PLDI):1463--1487. arXiv:2304.04812.
\item
  Manhaeve, R., Dumancic, S., Kimmig, A., Demeester, T., \& De
  Raedt, L. (2018). DeepProbLog: Neural Probabilistic Logic
  Programming. \emph{NeurIPS}.
\item
  Serafini, L. \& Garcez, A. d. (2016). Logic Tensor Networks: Deep
  Learning and Logical Reasoning from Data and Knowledge. \emph{NeSy
  Workshop}.
\item
  van Krieken, E., Acar, E., \& van Harmelen, F. (2022).
  Analyzing Differentiable Fuzzy Logic Operators. \emph{Artificial
  Intelligence} 302:103602. The differentiable-fuzzy-logic survey
  cited in §1.1-1; analyzes t-norm-derived AND/OR/IMPLIES
  operators in the neural-symbolic context and is the closest
  prior literature to Sutra\textquotesingle s polynomial approach.
\item
  Vergés, P., Heddes, M., Nunes, I., Givargis, T., \& Nicolau, A.
  (2023). HDCC: A Hyperdimensional Computing compiler for
  classification on embedded systems and high-performance
  computing. arXiv:2304.12398.
\item
  Yang, Z., Ishay, A., \& Lee, J. (2020). NeurASP: Embracing Neural
  Networks into Answer Set Programming. \emph{IJCAI}.
\item
  Plate, T. A. (1995). Holographic reduced representations. \emph{IEEE
  Transactions on Neural Networks} 6(3):623--641.
\item
  Shaw, et al. (2025). Developing a Foundation of Vector Symbolic
  Architectures Using Category Theory. arXiv:2501.05368. Cited in
  Related Work for independently identifying the construction of a
  purely functional VSA as a significant open direction.
\item
  Siegelmann, H. T. \& Sontag, E. D. (1992). On the computational
  power of neural nets. \emph{COLT \textquotesingle92}. Establishes that recurrent
  neural networks with rational weights are Turing-complete;
  Sutra\textquotesingle s tail-recursive loops over a fixed-width state vector
  take that recurrent form.
\item
  Smolensky, P. (1990). Tensor product variable binding and the
  representation of symbolic structures in connectionist systems.
  \emph{Artificial Intelligence} 46(1--2):159--216.
\item
  Ethayarajh, K. (2019). How Contextual are Contextualized Word
  Representations? Comparing the Geometry of BERT, ELMo, and GPT-2
  Embeddings. \emph{EMNLP-IJCNLP}. arXiv:1909.00512. The anisotropy /
  narrow-cone result: contextual embeddings occupy a narrow cone,
  inflating cosine similarity between unrelated items --- the
  precise reason frozen LLM embeddings are not the i.i.d.
  distribution textbook VSA assumes (§1, §2).
\item
  Gao, J., He, D., Tan, X., Qin, T., Wang, L., \& Liu, T.-Y.
  (2019). Representation Degeneration Problem in Training Natural
  Language Generation Models. \emph{ICLR}. arXiv:1907.12009.
\item
  Mu, J., Bhat, S., \& Viswanath, P. (2018). All-but-the-Top:
  Simple and Effective Postprocessing for Word Representations.
  \emph{ICLR}. arXiv:1702.01417.
\item
  Kingma, D. P. \& Ba, J. (2015). Adam: A Method for Stochastic
  Optimization. \emph{3rd International Conference on Learning
  Representations (ICLR)}. arXiv:1412.6980. The optimizer (run with
  its default \(\beta_1, \beta_2, \varepsilon\)) used for the §3.6
  differentiable-training experiment.
\item
  Lin, Z., Akin, H., Rao, R., Hie, B., Zhu, Z., Lu, W., Smetanin,
  N., Verkuil, R., Kabeli, O., Shmueli, Y., dos Santos Costa, A.,
  Fazel-Zarandi, M., Sercu, T., Candido, S., \& Rives, A. (2023).
  Evolutionary-scale prediction of atomic-level protein structure
  with a language model. \emph{Science} 379(6637):1123--1130. The ESM-2
  protein language model used as the non-text substrate in §3.2.
\item
  Zadeh, L. A. (1965). Fuzzy sets. \emph{Information and Control}
  8(3):338--353.
\item
  Jang, J.-S. R. (1993). ANFIS: Adaptive-Network-Based Fuzzy
  Inference System. \emph{IEEE Transactions on Systems, Man, and
  Cybernetics} 23(3):665--685.
\item
  Buckley, J. J. \& Hayashi, Y. (1994). Fuzzy neural networks: A
  survey. \emph{Fuzzy Sets and Systems} 66(1):1--13.
\end{itemize}

\begin{center}\rule{0.5\linewidth}{0.5pt}\end{center}

\newpage

\hypertarget{appendix}{%
\section*{Appendix}\label{appendix}}
\addcontentsline{toc}{section}{Appendix}

\hypertarget{appendix-a.-notation-extended-layout-and-primitive-operations}{%
\subsection*{Appendix A. Notation: extended layout and primitive operations}\label{appendix-a.-notation-extended-layout-and-primitive-operations}}
\addcontentsline{toc}{subsection}{Appendix A. Notation: extended layout and primitive operations}

We work in a fixed-dimensional real vector space \(\mathbb{R}^d\)
where \(d\) is the substrate\textquotesingle s embedding dimension (768 for
nomic-embed-text, 384 for all-minilm, 1024 for mxbai-embed-large,
320 for ESM-2). Every Sutra value carries the extended layout
\([\,\text{semantic}\mid\text{synthetic}\,]\), a \(d\)-dimensional
semantic block holding the substrate embedding, concatenated with
a small fixed-width synthetic block reserving canonical axes for
primitive types (real, imag, truth, char, loop-done) and slot
machinery (§3.3). Where notation does not distinguish, "vector"
means "the full extended-layout tensor."

The seven primitive operations are:

\begin{align*}
\mathrm{bind}(r, f)        &\;=\; R_r \, f, \qquad R_r = \mathrm{QR}\!\left(\mathrm{seed}=\mathrm{hash}(r)\right)\!.Q \\
\mathrm{unbind}(r, v)      &\;=\; R_r^{\!\top} v \\
\mathrm{bundle}(x, y)      &\;=\; \frac{x + y}{\lVert x + y \rVert + \varepsilon} \\
\mathrm{similarity}(x, y)  &\;=\; \frac{x \cdot y}{\lVert x \rVert \, \lVert y \rVert + \varepsilon} \\
\mathrm{normalize}(v)      &\;=\; \frac{v}{\lVert v \rVert + \varepsilon}
\end{align*}

plus the Lagrange Kleene gates (scalar \(\to\) scalar, exact on the
\(\{-1,0,+1\}^2\) grid, §1.1‑1) and the soft-halt cell
(state, halt \(\to\) state\('\), halt\('\), §3.4).

The Lagrange gates in closed form:

\begin{align*}
\mathrm{AND}(a, b)  &\;=\; \tfrac{1}{2}\!\left(a + b + ab - a^2 - b^2 + a^2 b^2\right) \\
\mathrm{NAND}(a, b) &\;=\; \tfrac{1}{2}\!\left(-a - b - ab + a^2 + b^2 - a^2 b^2\right) \\
\mathrm{OR}(a, b)   &\;=\; \tfrac{1}{2}\!\left(a + b - ab + a^2 + b^2 - a^2 b^2\right) \\
\mathrm{NOR}(a, b)  &\;=\; \tfrac{1}{2}\!\left(-a - b + ab - a^2 - b^2 + a^2 b^2\right) \\
\mathrm{NOT}(a)     &\;=\; -a \\
\mathrm{XOR}(a, b)  &\;=\; -ab \\
\mathrm{XNOR}(a, b) &\;=\; ab
\end{align*}

The soft-halt cell update is, in compact form,

\begin{align*}
s_{t+1}      &\;=\; R \, s_t                                && \text{(rotation step)} \\
h_t          &\;=\; \mathrm{Heaviside}\!\left(\mathrm{cond}(s_t)\right) && \text{(per-tick halt signal)} \\
H_t          &\;=\; \mathrm{sat}_{[0,1]}\!\left(\textstyle\sum_{k\le t} h_k\right) && \text{(cumulative monotone halt)} \\
\hat{s}_{t+1}&\;=\; H_t \, s_t + (1 - H_t)\, s_{t+1}        && \text{(soft-mux freeze)}
\end{align*}

Every right-hand side is a tensor expression with no Python
control flow. The compile-time primitives \texttt{RotationFor} and
\texttt{embed} produce constants \(R_r\) and basis vectors at compile
time and are not part of the runtime tensor graph.

\hypertarget{appendix-b.-extended-state-vector-layout-per-axis-assignments}{%
\subsection*{Appendix B. Extended-state-vector layout: per-axis assignments}\label{appendix-b.-extended-state-vector-layout-per-axis-assignments}}
\addcontentsline{toc}{subsection}{Appendix B. Extended-state-vector layout: per-axis assignments}

§3.3 describes the \texttt{{[}semantic\ \textbar{}\ synthetic{]}} layout in prose. The
diagram and per-axis purpose table below give the concrete
allocation referenced in \texttt{codegen\_pytorch.py}:

\begin{verbatim}
          +-------------------------+----+----+----+----+----+----------+
   value  | semantic block          | R  | I  | T  | C  | L  | slots... |
          +-------------------------+----+----+----+----+----+----------+
          |<-- semantic_dim ------->|<--- synthetic_dim ----------------|>
                                       0    1    2    3    4    5..
                                      REAL IMAG TRUTH CHAR LOOP_DONE
                                                      _FLAG
\end{verbatim}

\begin{longtable}[]{@{}
  >{\raggedright\arraybackslash}p{(\columnwidth - 2\tabcolsep) * \real{0.2375}}
  >{\raggedright\arraybackslash}p{(\columnwidth - 2\tabcolsep) * \real{0.7625}}@{}}
\toprule\noalign{}
\begin{minipage}[b]{\linewidth}\raggedright
Index
\end{minipage} & \begin{minipage}[b]{\linewidth}\raggedright
Purpose
\end{minipage} \\
\midrule\noalign{}
\endhead
\bottomrule\noalign{}
\endlastfoot
\texttt{synthetic{[}0{]}} & \texttt{AXIS\_REAL} (real component for int/float/complex) \\
\texttt{synthetic{[}1{]}} & \texttt{AXIS\_IMAG} (imaginary component for complex) \\
\texttt{synthetic{[}2{]}} & \texttt{AXIS\_TRUTH} (fuzzy truth scalar; bool/comparisons) \\
\texttt{synthetic{[}3{]}} & \texttt{AXIS\_CHAR\_FLAG} (marks char primitives) \\
\texttt{synthetic{[}4{]}} & \texttt{AXIS\_LOOP\_DONE} (substrate-side completion flag) \\
\texttt{synthetic{[}5..{]}} & \texttt{SLOT\_BASE}: disjoint 2D Givens slots for variable storage \\
\end{longtable}

At \texttt{semantic\_dim\ =\ 768} (nomic-embed-text), \texttt{synthetic\_dim\ =\ 100}
accommodates the five canonical axes plus 47 disjoint Givens slots.

\hypertarget{appendix-c.-capacity-full-per-substrate-sweeps}{%
\subsection*{Appendix C. Capacity: full per-substrate sweeps}\label{appendix-c.-capacity-full-per-substrate-sweeps}}
\addcontentsline{toc}{subsection}{Appendix C. Capacity: full per-substrate sweeps}

Cross-substrate decode accuracy at full bundle widths
k ∈ \{2, 4, 8, 16, 24, 32, 48\}. The four substrates use 84-entry
vocabularies (LLM substrates: 84-word noun set spanning animals,
foods, objects, places, abstract nouns; ESM-2: 84-sequence
amino-acid set covering canonical signal peptides,
cell-penetrating peptides, antimicrobial peptides, classic
affinity-tag motifs, and deterministic random k-mers). All
embeddings are unit-normalized; nomic-embed-text and ESM-2 are
additionally mean-centered.

\textbf{nomic-embed-text (768-d, mean-centered):}

\begin{longtable}[]{@{}
  >{\raggedleft\arraybackslash}p{(\columnwidth - 8\tabcolsep) * \real{0.2000}}
  >{\raggedleft\arraybackslash}p{(\columnwidth - 8\tabcolsep) * \real{0.2000}}
  >{\raggedleft\arraybackslash}p{(\columnwidth - 8\tabcolsep) * \real{0.2000}}
  >{\raggedleft\arraybackslash}p{(\columnwidth - 8\tabcolsep) * \real{0.2000}}
  >{\raggedleft\arraybackslash}p{(\columnwidth - 8\tabcolsep) * \real{0.2000}}@{}}
\toprule\noalign{}
\begin{minipage}[b]{\linewidth}\raggedleft
k
\end{minipage} & \begin{minipage}[b]{\linewidth}\raggedleft
rotation accuracy
\end{minipage} & \begin{minipage}[b]{\linewidth}\raggedleft
rotation signal cos
\end{minipage} & \begin{minipage}[b]{\linewidth}\raggedleft
Hadamard accuracy
\end{minipage} & \begin{minipage}[b]{\linewidth}\raggedleft
Hadamard signal cos
\end{minipage} \\
\midrule\noalign{}
\endhead
\bottomrule\noalign{}
\endlastfoot
2 & 100.0\% & +0.703 & 95.0\% & +0.488 \\
4 & 100.0\% & +0.497 & 95.0\% & +0.400 \\
8 & 100.0\% & +0.354 & 87.5\% & +0.307 \\
16 & 100.0\% & +0.251 & 84.4\% & +0.230 \\
24 & 100.0\% & +0.203 & 60.8\% & +0.189 \\
32 & 99.1\% & +0.176 & 63.1\% & +0.167 \\
48 & 93.3\% & +0.144 & 48.3\% & +0.136 \\
\end{longtable}

\textbf{all-minilm (384-d):}

\begin{longtable}[]{@{}
  >{\raggedleft\arraybackslash}p{(\columnwidth - 8\tabcolsep) * \real{0.2000}}
  >{\raggedleft\arraybackslash}p{(\columnwidth - 8\tabcolsep) * \real{0.2000}}
  >{\raggedleft\arraybackslash}p{(\columnwidth - 8\tabcolsep) * \real{0.2000}}
  >{\raggedleft\arraybackslash}p{(\columnwidth - 8\tabcolsep) * \real{0.2000}}
  >{\raggedleft\arraybackslash}p{(\columnwidth - 8\tabcolsep) * \real{0.2000}}@{}}
\toprule\noalign{}
\begin{minipage}[b]{\linewidth}\raggedleft
k
\end{minipage} & \begin{minipage}[b]{\linewidth}\raggedleft
rotation accuracy
\end{minipage} & \begin{minipage}[b]{\linewidth}\raggedleft
rotation signal cos
\end{minipage} & \begin{minipage}[b]{\linewidth}\raggedleft
Hadamard accuracy
\end{minipage} & \begin{minipage}[b]{\linewidth}\raggedleft
Hadamard signal cos
\end{minipage} \\
\midrule\noalign{}
\endhead
\bottomrule\noalign{}
\endlastfoot
2 & 100.0\% & +0.711 & 45.0\% & +0.386 \\
4 & 100.0\% & +0.506 & 10.0\% & +0.335 \\
8 & 100.0\% & +0.356 & 7.5\% & +0.315 \\
16 & 92.5\% & +0.252 & 3.1\% & +0.299 \\
24 & 76.2\% & +0.203 & 2.9\% & +0.300 \\
32 & 66.9\% & +0.179 & 2.5\% & +0.297 \\
48 & 42.3\% & +0.144 & 1.7\% & +0.294 \\
\end{longtable}

\textbf{mxbai-embed-large (1024-d):}

\begin{longtable}[]{@{}
  >{\raggedleft\arraybackslash}p{(\columnwidth - 8\tabcolsep) * \real{0.2000}}
  >{\raggedleft\arraybackslash}p{(\columnwidth - 8\tabcolsep) * \real{0.2000}}
  >{\raggedleft\arraybackslash}p{(\columnwidth - 8\tabcolsep) * \real{0.2000}}
  >{\raggedleft\arraybackslash}p{(\columnwidth - 8\tabcolsep) * \real{0.2000}}
  >{\raggedleft\arraybackslash}p{(\columnwidth - 8\tabcolsep) * \real{0.2000}}@{}}
\toprule\noalign{}
\begin{minipage}[b]{\linewidth}\raggedleft
k
\end{minipage} & \begin{minipage}[b]{\linewidth}\raggedleft
rotation accuracy
\end{minipage} & \begin{minipage}[b]{\linewidth}\raggedleft
rotation signal cos
\end{minipage} & \begin{minipage}[b]{\linewidth}\raggedleft
Hadamard accuracy
\end{minipage} & \begin{minipage}[b]{\linewidth}\raggedleft
Hadamard signal cos
\end{minipage} \\
\midrule\noalign{}
\endhead
\bottomrule\noalign{}
\endlastfoot
2 & 100.0\% & +0.708 & 15.0\% & +0.311 \\
4 & 100.0\% & +0.500 & 2.5\% & +0.304 \\
8 & 100.0\% & +0.353 & 2.5\% & +0.295 \\
16 & 98.8\% & +0.251 & 1.2\% & +0.294 \\
24 & 95.8\% & +0.203 & 0.8\% & +0.293 \\
32 & 85.3\% & +0.176 & 0.9\% & +0.292 \\
48 & 72.1\% & +0.146 & 1.0\% & +0.291 \\
\end{longtable}

\textbf{ESM-2 small protein language model (320-d, mean-centered):}

\begin{longtable}[]{@{}
  >{\raggedleft\arraybackslash}p{(\columnwidth - 8\tabcolsep) * \real{0.2000}}
  >{\raggedleft\arraybackslash}p{(\columnwidth - 8\tabcolsep) * \real{0.2000}}
  >{\raggedleft\arraybackslash}p{(\columnwidth - 8\tabcolsep) * \real{0.2000}}
  >{\raggedleft\arraybackslash}p{(\columnwidth - 8\tabcolsep) * \real{0.2000}}
  >{\raggedleft\arraybackslash}p{(\columnwidth - 8\tabcolsep) * \real{0.2000}}@{}}
\toprule\noalign{}
\begin{minipage}[b]{\linewidth}\raggedleft
k
\end{minipage} & \begin{minipage}[b]{\linewidth}\raggedleft
rotation accuracy
\end{minipage} & \begin{minipage}[b]{\linewidth}\raggedleft
rotation signal cos
\end{minipage} & \begin{minipage}[b]{\linewidth}\raggedleft
Hadamard accuracy
\end{minipage} & \begin{minipage}[b]{\linewidth}\raggedleft
Hadamard signal cos
\end{minipage} \\
\midrule\noalign{}
\endhead
\bottomrule\noalign{}
\endlastfoot
2 & 100.0\% & +0.713 & 75.0\% & +0.470 \\
4 & 100.0\% & +0.501 & 50.0\% & +0.323 \\
8 & 100.0\% & +0.349 & 28.7\% & +0.257 \\
16 & 90.6\% & +0.252 & 16.2\% & +0.185 \\
24 & 77.1\% & +0.205 & 11.2\% & +0.171 \\
32 & 61.9\% & +0.174 & 6.2\% & +0.141 \\
48 & 44.2\% & +0.143 & 4.2\% & +0.117 \\
\end{longtable}

The signal cosine for Hadamard is comparable to rotation\textquotesingle s, but
the noise floor is much higher because the elementwise product
of correlated real-valued embeddings produces a result that
overlaps with many distractors in the codebook rather than
near-orthogonally with one.

\hypertarget{appendix-d.-crosstalk-depth-full-per-substrate-l-sweep}{%
\subsection*{Appendix D. Crosstalk depth: full per-substrate L-sweep}\label{appendix-d.-crosstalk-depth-full-per-substrate-l-sweep}}
\addcontentsline{toc}{subsection}{Appendix D. Crosstalk depth: full per-substrate L-sweep}

The §3.2.1 protocol: chain length L ∈ \{1, 2, 4, 8, 16, 32\}, 20
trials, bundle width 4 (3 distractors per cycle). Forward-bind
through L role rotations bundling 3 distractor (role, filler)
pairs at each step; unbind in reverse and decode. Two flavors:
\emph{raw} (no cleanup) and \emph{snap} (argmax-cosine cleanup against the
codebook after each unbind step).

\begin{longtable}[]{@{}
  >{\raggedright\arraybackslash}p{(\columnwidth - 12\tabcolsep) * \real{0.2500}}
  >{\raggedleft\arraybackslash}p{(\columnwidth - 12\tabcolsep) * \real{0.1184}}
  >{\raggedleft\arraybackslash}p{(\columnwidth - 12\tabcolsep) * \real{0.1184}}
  >{\raggedleft\arraybackslash}p{(\columnwidth - 12\tabcolsep) * \real{0.1184}}
  >{\raggedleft\arraybackslash}p{(\columnwidth - 12\tabcolsep) * \real{0.1316}}
  >{\raggedleft\arraybackslash}p{(\columnwidth - 12\tabcolsep) * \real{0.1316}}
  >{\raggedleft\arraybackslash}p{(\columnwidth - 12\tabcolsep) * \real{0.1316}}@{}}
\toprule\noalign{}
\begin{minipage}[b]{\linewidth}\raggedright
substrate
\end{minipage} & \begin{minipage}[b]{\linewidth}\raggedleft
L=1 raw
\end{minipage} & \begin{minipage}[b]{\linewidth}\raggedleft
L=2 raw
\end{minipage} & \begin{minipage}[b]{\linewidth}\raggedleft
L=4 raw
\end{minipage} & \begin{minipage}[b]{\linewidth}\raggedleft
L=1 snap
\end{minipage} & \begin{minipage}[b]{\linewidth}\raggedleft
L=2 snap
\end{minipage} & \begin{minipage}[b]{\linewidth}\raggedleft
L=4 snap
\end{minipage} \\
\midrule\noalign{}
\endhead
\bottomrule\noalign{}
\endlastfoot
nomic-embed-text & 100\% & 100\% & 20\% & 100\% & 10\% & 0\% \\
all-minilm & 100\% & 100\% & 5\% & 100\% & 0\% & 0\% \\
mxbai-embed-large & 100\% & 100\% & 5\% & 100\% & 0\% & 0\% \\
\end{longtable}

By chain length 8 raw accuracy is at chance (1/84) on all three
substrates. Snap is \emph{worse} than raw past chain length 1: a
hard codebook commitment converts soft noise into a
high-confidence wrong answer that the next unbind cannot
recover from. The runtime does not implicitly snap between
operations; cleanup is an explicit step the program schedules
where it knows the codebook is the right reference. Reproduction
script: \texttt{experiments/crosstalk\_chain.py}; raw JSON in
\texttt{experiments/crosstalk\_chain\_results.json}.

\hypertarget{appendix-e.-codebook-implementation-details}{%
\subsection*{Appendix E. Codebook implementation details}\label{appendix-e.-codebook-implementation-details}}
\addcontentsline{toc}{subsection}{Appendix E. Codebook implementation details}

The §3.5 codebook is implemented as an embedded vector database
(internally SutraDB) shipped as part of the compiler, analogous
to SQLite being embedded in an application rather than run as a
separate service. The data model is RDF
triples with f32-vector literals as the object position, indexed
by a built-in HNSW index for nearest-neighbor decode. The
on-disk format is a \texttt{.sdb} file that travels alongside the
compiled Python module; no external service, no separate
install, no network dependency. Every embedded string in a
Sutra program is inserted with the embedding as the object of a
triple typed \texttt{\textless{}http://sutra.dev/f32vec\textgreater{}}. Strings declared but
unused in expressions are still inserted, so they remain
decodable. The compiled module\textquotesingle s Python data section never
carries the embeddings, they live in the \texttt{.sdb} file, an
artifact of compilation, not a service the runtime contacts.

\texttt{nearest\_string} runs over an HNSW (Hierarchical Navigable
Small World) approximate-nearest-neighbor graph maintained by
the triplestore. HNSW (Malkov \& Yashunin, TPAMI 2020) has
\textbf{O(log N) expected and worst-case query time} under standard
graph-construction parameters; it has displaced linear scan as
the default ANN index in Faiss, Milvus, Weaviate, Qdrant, and
most production vector databases. A 100-string codebook and a
100,000-string codebook have comparable decode latency at
runtime, modulo HNSW\textquotesingle s tunable \texttt{M} (graph degree) and
\texttt{ef\_search} (beam width); the cost difference is roughly one
extra graph hop per 10× growth in N.

\hypertarget{appendix-f.-worked-lowering-of-a-two-field-bundled-record}{%
\subsection*{Appendix F. Worked lowering of a two-field bundled record}\label{appendix-f.-worked-lowering-of-a-two-field-bundled-record}}
\addcontentsline{toc}{subsection}{Appendix F. Worked lowering of a two-field bundled record}

The body §4.2 sketches the lowering of
\(\mathrm{encode2}(r_a, f_a, r_b, f_b) \,:=\, \mathrm{bundle}(\mathrm{bind}(r_a, f_a),\,\mathrm{bind}(r_b, f_b))\).
Here we trace each stage with the explicit residual.

\textbf{Stage 1: AST after parse.} A tree of \texttt{Call} nodes over named
identifiers: \texttt{Call("bundle",\ Call("bind",\ r\_a,\ f\_a),\ Call("bind",\ r\_b,\ f\_b))}.

\textbf{Stage 2: beta reduction by stdlib inlining.} \texttt{bind},
\texttt{bundle}, and \texttt{normalize} are stdlib functions:
\(\mathrm{bind}(r, f) \equiv \mathrm{RotationFor}(r)\,f\),
\(\mathrm{bundle}(x, y) \equiv \mathrm{normalize}(x + y)\),
\(\mathrm{normalize}(v) \equiv v / (\lVert v\rVert + \varepsilon)\).
After substitution the body becomes

\[
\mathrm{normalize}\!\bigl(\mathrm{RotationFor}(r_a)\,f_a \;+\; \mathrm{RotationFor}(r_b)\,f_b\bigr).
\]

No \texttt{bind} or \texttt{bundle} symbol remains; the residual is straight-
line algebra over four tensor primitives.

\textbf{Stage 3: compile-time constant resolution.}
\(\mathrm{RotationFor}(r)\) is a compile-time function returning
\(R = \mathrm{QR}(\mathrm{seed}=\mathrm{hash}(r)).Q\). The compiler
evaluates it for each role at compile time, freezes the results
as constant tensors \(R_a\) and \(R_b\), and stores them in the
rotation cache. The body becomes
\(\mathrm{normalize}(R_a\,f_a + R_b\,f_b)\), \(R_a\) and \(R_b\) are
now load-bearing constants in the same sense as the weight
matrices of a feed-forward network.

\textbf{Stage 4: peephole fusion.} The simplifier recognizes
\(\mathrm{normalize}\!\bigl(\textstyle\sum_i R_i\,f_i\bigr)\) as the
bundle-of-binds pattern and rewrites it to
\texttt{\_VSA.bundle\_of\_binds({[}(R\_a,\ f\_a),\ (R\_b,\ f\_b){]})}, one kernel
launch instead of two matmuls + add + norm.

\textbf{Stage 5: leaf tensor ops at runtime.} \texttt{bundle\_of\_binds}
stacks rotations into a \((k, d, d)\) tensor, stacks fillers into
\((k, d)\), runs one batched einsum + sum + L2-normalize:

\begin{align*}
v          &\;=\; \sum_{k} R_k\,f_k \;=\; \mathtt{einsum("kij,kj->i",\; \mathrm{stack}([R_a, R_b]),\; \mathrm{stack}([f_a, f_b]))} \\
\mathrm{encode2} &\;=\; v \,/\, (\lVert v\rVert + \varepsilon)
\end{align*}

The compiled forward pass for \texttt{encode2} is exactly those three
torch calls (einsum, linalg.norm, divide) over precomputed
\(R_a, R_b\) and runtime-supplied \(f_a, f_b\).

\hypertarget{appendix-g.-3.6-differentiable-training-vocabulary}{%
\subsection*{Appendix G. §3.6 differentiable-training vocabulary}\label{appendix-g.-3.6-differentiable-training-vocabulary}}
\addcontentsline{toc}{subsection}{Appendix G. §3.6 differentiable-training vocabulary}

Twenty categories of fifty words each (992 unique after
deduplication), embedded via nomic-embed-text:

\begin{itemize}
\tightlist
\item
  \textbf{animal}: dog, cat, bird, fish, horse, lion, tiger, elephant, rabbit, monkey, bear, wolf, fox, deer, mouse, snake, frog, turtle, dolphin, whale, shark, eagle, owl, sparrow, crow, robin, parrot, swan, duck, goose, chicken, cow, pig, sheep, goat, donkey, camel, giraffe, kangaroo, koala, panda, leopard, cheetah, hippopotamus, rhinoceros, antelope, buffalo, hedgehog, squirrel, raccoon
\item
  \textbf{vehicle}: car, truck, airplane, boat, bicycle, motorcycle, bus, train, ship, helicopter, tractor, scooter, van, taxi, jeep, sailboat, kayak, canoe, raft, submarine, glider, jet, rocket, spaceship, sled, skateboard, wagon, carriage, chariot, ambulance, firetruck, limousine, minivan, hatchback, sedan, coupe, convertible, pickup, trailer, ferry, yacht, dinghy, blimp, balloon, hovercraft, tram, moped, tricycle, rollerblade, unicycle
\item
  \textbf{food}: apple, bread, cheese, rice, pasta, banana, salad, soup, meat, pizza, sandwich, burger, taco, sushi, cake, cookie, pie, donut, muffin, pancake, waffle, bagel, croissant, omelet, salmon, tuna, beef, pork, lamb, bacon, ham, sausage, steak, lobster, shrimp, crab, oyster, clam, broccoli, carrot, lettuce, tomato, potato, cucumber, onion, garlic, pepper, eggplant, spinach, mushroom
\item
  \textbf{color}: red, blue, green, yellow, orange, purple, black, white, brown, pink, gray, cyan, magenta, violet, indigo, turquoise, teal, lavender, maroon, crimson, scarlet, ruby, gold, silver, bronze, copper, beige, tan, ivory, charcoal, navy, sapphire, emerald, jade, olive, lime, mint, coral, peach, plum, mauve, fuchsia, amber, ochre, sienna, mahogany, chocolate, caramel, mustard, azure
\item
  \textbf{clothing}: shirt, pants, dress, hat, shoes, jacket, socks, gloves, scarf, belt, sweater, hoodie, jeans, shorts, skirt, blouse, coat, cap, beanie, mittens, tights, leggings, vest, blazer, suit, tuxedo, gown, robe, kimono, kilt, poncho, cloak, cape, sneakers, boots, sandals, slippers, heels, loafers, tie, bowtie, cufflinks, watch, ring, necklace, earrings, bracelet, anklet, brooch, headband
\item
  \textbf{weather}: rain, snow, wind, cloud, storm, fog, frost, hail, thunder, lightning, drizzle, downpour, blizzard, hurricane, tornado, cyclone, typhoon, sleet, mist, haze, smog, sunshine, sunlight, sunset, sunrise, dawn, dusk, twilight, breeze, gust, gale, humidity, drought, flood, monsoon, snowfall, snowstorm, rainstorm, sandstorm, heatwave, chill, dew, hailstorm, thaw, overcast, sunny, cloudy, rainy, snowy, windy
\item
  \textbf{emotion}: joy, sadness, anger, fear, love, hope, surprise, disgust, pride, envy, happiness, grief, rage, anxiety, affection, despair, delight, shame, guilt, confidence, contentment, jealousy, regret, sorrow, frustration, satisfaction, awe, wonder, gratitude, compassion, sympathy, empathy, irritation, boredom, excitement, enthusiasm, calm, serenity, melancholy, nostalgia, longing, embarrassment, humiliation, indifference, ecstasy, bliss, dread, terror, amusement, loneliness
\item
  \textbf{tool}: hammer, saw, drill, wrench, screwdriver, knife, scissors, pliers, axe, shovel, rake, hoe, spade, pickaxe, crowbar, mallet, chisel, sander, level, ruler, vise, clamp, ratchet, socket, awl, scraper, trowel, broom, mop, sponge, bucket, ladder, jackhammer, sledgehammer, paintbrush, roller, stapler, tongs, tweezers, calipers, magnifier, flashlight, multimeter, wirecutter, hacksaw, router, torch, soldering\_iron, drillbit, screwbit
\item
  \textbf{instrument}: guitar, piano, drum, violin, flute, trumpet, saxophone, harp, cello, clarinet, banjo, mandolin, ukulele, harmonica, accordion, organ, keyboard, synthesizer, xylophone, tambourine, maracas, bongos, marimba, vibraphone, glockenspiel, bagpipes, oboe, bassoon, trombone, tuba, lute, sitar, koto, zither, dulcimer, cymbal, gong, triangle, cowbell, snare, kettledrum, recorder, piccolo, fife, didgeridoo, theremin, viola, double\_bass, fiddle, ocarina
\item
  \textbf{profession}: doctor, teacher, lawyer, engineer, nurse, chef, artist, scientist, farmer, plumber, electrician, carpenter, mechanic, pilot, sailor, soldier, judge, journalist, writer, poet, painter, sculptor, musician, actor, dancer, singer, photographer, architect, dentist, surgeon, pharmacist, veterinarian, librarian, accountant, banker, broker, programmer, designer, manager, secretary, butcher, baker, gardener, tailor, jeweler, barber, chemist, biologist, physicist, mathematician
\item
  \textbf{body\_part}: head, hand, foot, eye, ear, nose, mouth, leg, arm, finger, toe, knee, elbow, shoulder, hip, neck, back, chest, stomach, heart, brain, lung, liver, kidney, bone, muscle, skin, hair, throat, jaw, chin, cheek, forehead, eyebrow, eyelash, lip, tongue, palm, wrist, ankle, thumb, heel, spine, rib, scalp, nostril, gum, knuckle, tendon, vein
\item
  \textbf{plant}: tree, flower, grass, bush, vine, fern, moss, herb, weed, leaf, stem, branch, bark, blossom, petal, oak, maple, willow, birch, cedar, bamboo, cactus, rose, tulip, daisy, lily, sunflower, orchid, ivy, basil, rosemary, thyme, sage, lavender, dandelion, clover, lotus, magnolia, sycamore, redwood, baobab, eucalyptus, juniper, hemlock, fir, spruce, ash, elm, poplar, chestnut
\item
  \textbf{furniture}: chair, table, sofa, bed, desk, shelf, drawer, cabinet, wardrobe, dresser, nightstand, ottoman, bench, stool, recliner, futon, couch, armchair, bookcase, sideboard, buffet, cupboard, hutch, vanity, headboard, footboard, mattress, pillow, cushion, blanket, quilt, comforter, lamp, mirror, rug, carpet, curtain, blind, shutter, hammock, cradle, crib, bassinet, highchair, rocker, loveseat, settee, divan, chaise, headrest
\item
  \textbf{building}: house, apartment, mansion, cottage, cabin, hut, igloo, tent, palace, castle, fortress, tower, skyscraper, office, factory, warehouse, store, mall, restaurant, hotel, motel, hospital, school, university, library, museum, theater, stadium, arena, church, temple, mosque, synagogue, cathedral, chapel, monastery, abbey, barn, shed, garage, basement, attic, cellar, lobby, lounge, hallway, corridor, atrium, foyer, balcony
\item
  \textbf{country}: France, Germany, Italy, Spain, Portugal, England, Scotland, Ireland, Norway, Sweden, Finland, Denmark, Iceland, Russia, Poland, Greece, Turkey, Egypt, Morocco, Algeria, Kenya, Nigeria, Ethiopia, Ghana, Senegal, Mali, Sudan, Uganda, Tanzania, Madagascar, China, Japan, Korea, Vietnam, Thailand, Malaysia, Indonesia, India, Pakistan, Bangladesh, Iran, Iraq, Israel, Lebanon, Australia, Canada, Mexico, Brazil, Argentina, Chile
\item
  \textbf{sport}: football, basketball, baseball, soccer, tennis, golf, hockey, rugby, cricket, volleyball, swimming, running, cycling, skiing, snowboarding, surfing, sailing, rowing, kayaking, climbing, hiking, boxing, wrestling, fencing, archery, shooting, fishing, hunting, polo, badminton, ping\_pong, squash, racquetball, lacrosse, handball, dodgeball, kickball, gymnastics, diving, weightlifting, judo, karate, taekwondo, sumo, marathon, triathlon, decathlon, biathlon, skating, bowling
\item
  \textbf{drink}: water, juice, milk, tea, coffee, soda, beer, wine, whiskey, vodka, rum, gin, tequila, brandy, cognac, champagne, cocktail, smoothie, milkshake, lemonade, cider, ale, lager, stout, bourbon, scotch, sake, mead, punch, eggnog, kombucha, kefir, espresso, latte, cappuccino, mocha, americano, macchiato, frappe, hot\_chocolate, cordial, shake, slushie, syrup, fizz, brew, tonic, infusion, ginger\_ale, root\_beer
\item
  \textbf{metal}: gold, silver, copper, iron, steel, aluminum, brass, bronze, tin, lead, zinc, nickel, platinum, titanium, chromium, mercury, magnesium, lithium, sodium, potassium, calcium, uranium, plutonium, palladium, tungsten, vanadium, cobalt, manganese, beryllium, gallium, indium, antimony, bismuth, cadmium, cerium, neodymium, osmium, rhodium, ruthenium, tantalum, thallium, thorium, yttrium, scandium, hafnium, niobium, molybdenum, rhenium, iridium, rubidium
\item
  \textbf{shape}: circle, square, triangle, rectangle, oval, ellipse, pentagon, hexagon, octagon, diamond, rhombus, trapezoid, parallelogram, polygon, sphere, cube, cylinder, cone, pyramid, prism, cuboid, tetrahedron, dodecahedron, icosahedron, octahedron, torus, helix, spiral, crescent, star, heart, arrow, cross, line, curve, arc, ring, loop, knot, dot, vertex, edge, angle, parabola, hyperbola, sine, wave, zigzag, scallop, annulus
\item
  \textbf{fabric}: cotton, wool, silk, linen, polyester, nylon, denim, leather, suede, velvet, satin, lace, tweed, cashmere, mohair, fleece, fur, canvas, burlap, jute, flannel, chiffon, organza, taffeta, brocade, damask, paisley, gingham, plaid, herringbone, corduroy, microfiber, spandex, lycra, rayon, viscose, acrylic, polypropylene, jersey, knit, sherpa, gabardine, twill, muslin, gauze, mesh, vinyl, tulle, georgette, voile
\end{itemize}

\hypertarget{appendix-h.-reproduction-details-and-hyperparameters}{%
\subsection*{Appendix H. Reproduction details and hyperparameters}\label{appendix-h.-reproduction-details-and-hyperparameters}}
\addcontentsline{toc}{subsection}{Appendix H. Reproduction details and hyperparameters}

Per-experiment configuration. All scripts live under \texttt{experiments/}
in the source repository; each writes a JSON results file to the
same directory on completion. RNG seeds are fixed in the source
files cited; re-running reproduces the numbers reported in the
body to the precision reported.

\begin{itemize}
\tightlist
\item
  \textbf{Rotation vs Hadamard, LLM} (§3.2) --- script
  \texttt{rotation\_binding\_capacity\_llm.py}; 10 trials per \emph{k}; substrates
  nomic-embed-text, all-minilm, mxbai-embed-large; seeds fixed per
  script.
\item
  \textbf{Rotation vs Hadamard, ESM-2} (§3.2) --- script
  \texttt{rotation\_binding\_capacity\_bioinformatics.py}; 10 trials per \emph{k};
  substrate \texttt{facebook/esm2\_t6\_8M\_UR50D}; seeds 1729 and 2718.
\item
  \textbf{Crosstalk depth} (§3.2.1) --- script \texttt{crosstalk\_chain.py}; 20
  trials per chain length \emph{L}; three LLM substrates; seeds fixed per
  script.
\item
  \textbf{Differentiable training} (§3.6) --- script
  \texttt{differentiable\_training\_compiled.py\ -\/-batched}; 3 seeds × 30 epochs
  (K=5, 50 words); substrate nomic-embed-text (frozen); optimizer
  Adam, lr=0.01; seeds 0--2.
\item
  \textbf{Trained weight → legible source} (§3.7) --- script
  \texttt{differentiable\_training\_weighted.py}; 2 seeds × 30 epochs (K=3, 24
  words); substrate nomic-embed-text (frozen); optimizer Adam,
  lr=0.02; seeds 0--1.
\end{itemize}

The differentiable-training run (\texttt{differentiable\_training\_compiled.py})
generates a \texttt{.su} fuzzy-rule classifier, compiles it with the
PyTorch codegen, and backpropagates five randomly-initialized
unit-normalized prototype vectors through the \emph{emitted} \texttt{rule}
function (the compiler\textquotesingle s \texttt{\_VSA.similarity} composed with the
Lagrange--Kleene polynomials). Five classes, ten words each
(50 inputs); embeddings are precomputed once and cached to
\texttt{.diff\_train\_embeddings.pt} so reruns skip the embed step. A
build-time assertion checks the emitted \texttt{similarity} is not
\texttt{float()}-collapsed (Stage A0), so gradients provably flow through
the compiled graph rather than a reimplementation. The forward is
evaluated batched via \texttt{torch.vmap} over that same emitted \texttt{rule};
a runtime assertion requires the batched and per-sample
evaluations to agree within 10⁻⁴ before training, so the fast
path is the identical compiled computation, not a substitute.

The §3.7 run (\texttt{differentiable\_training\_weighted.py}) adds a
\texttt{number\ w} parameter to the \texttt{.su} rule, trains \(w\) together with
the prototypes through the same emitted graph, then regenerates
the rule with \(w^{*}\) substituted as a numeric literal and the
parameter removed, recompiles that source through the PyTorch
codegen, and asserts the recompiled program\textquotesingle s logits match the
parametric model (max per-logit difference \textless{} 10⁻⁴) --- a
source-level training round-trip, not a benchmark.

Hardware used for the numbers in the body: CPU torch on a single
laptop (no CUDA). The §3.6 compiled run takes ≈ 230 s (K=5, 50 words, 30 epochs, 3 seeds) via the batched \texttt{torch.vmap} path over the same emitted ops; the equivalent per-sample Python driver produces the bit-identical result but takes ≈ 6.2 h --- an interpreter-overhead artifact, not a cost of the compiled graph;
the §3.7 weighted round-trip (K=3, 24 words, 30 epochs, 2 seeds, per-sample --- small enough that the driver cost is ≈ 2.5 min) completes including the recompile check;
the §3.2 capacity sweeps complete in \textasciitilde2 min per substrate; the
§3.2.1 crosstalk sweep completes in \textasciitilde5 min. Re-running on CUDA
should reproduce the same accuracy numbers since the operations
are deterministic given a seed.

\hypertarget{appendix-i.-demonstration-corpus}{%
\subsection*{Appendix I. Demonstration corpus}\label{appendix-i.-demonstration-corpus}}
\addcontentsline{toc}{subsection}{Appendix I. Demonstration corpus}

The smoke test (\texttt{examples/\_smoke\_test.py}) compiles and runs ten
\texttt{.su} programs end-to-end and asserts each output against a
hardcoded expected value. The programs collectively exercise the
language features the body claims, with no Python control flow on
the runtime path:

\begin{longtable}[]{@{}
  >{\raggedright\arraybackslash}p{(\columnwidth - 2\tabcolsep) * \real{0.5000}}
  >{\raggedright\arraybackslash}p{(\columnwidth - 2\tabcolsep) * \real{0.5000}}@{}}
\toprule\noalign{}
\begin{minipage}[b]{\linewidth}\raggedright
Program
\end{minipage} & \begin{minipage}[b]{\linewidth}\raggedright
Feature exercised
\end{minipage} \\
\midrule\noalign{}
\endhead
\bottomrule\noalign{}
\endlastfoot
\texttt{hello\_world.su} & embed + retrieve (minimal program) \\
\texttt{fuzzy\_branching.su} & weighted-superposition conditional \\
\texttt{role\_filler\_record.su} & bind / bundle / unbind on a 3-field record (§2.1) \\
\texttt{classifier.su} & cosine-similarity classifier over a small codebook \\
\texttt{analogy.su} & associative pair memory: capital → country recovery via unbind \\
\texttt{knowledge\_graph.su} & (subject, relation, object) triple encode + decode \\
\texttt{predicate\_lookup.su} & bind-keyed dictionary read \\
\texttt{fuzzy\_dispatch.su} & Lagrange-Kleene-gated dispatch among handlers \\
\texttt{nearest\_phrase.su} & top-1 phrase retrieval over a \texttt{.sdb} codebook \\
\texttt{sequence.su} & foreach reduction over a list \\
\end{longtable}

Loop coverage lives in \texttt{examples/do\_while\_adder.su} and the
23-case \texttt{tests/test\_loop\_function\_decl.py} suite. The §3.6
differentiable-training experiment uses the same primitive set
the smoke-test programs are built from, no Sutra-runtime
extensions, just compilation of \texttt{.su} source to PyTorch
tensor ops.

\end{document}